%% file: main.tex
\documentclass[twoside,11pt]{article}

\pdfoutput=1
\usepackage{jmlr2e}
\usepackage[utf8]{inputenc}
\usepackage{amsmath}
\usepackage{amsfonts}
\usepackage{graphicx}
\usepackage{wrapfig}
\usepackage{float}
\usepackage[font={small}]{caption}
\usepackage[colorinlistoftodos, textsize=tiny]{todonotes}
\usepackage{mathtools}
\usepackage{color}
\usepackage{natbib}
\usepackage{verbatim}
\usepackage{tikz}
\usepackage{enumitem}  % for tighter list spacing
\usepackage{cancel}
\usepackage{subcaption}
\usepackage{array,multirow}
\usepackage{booktabs}
\newcommand{\ra}[1]{\renewcommand{\arraystretch}{#1}}

\DeclareMathOperator*{\diag}{diag}

\definecolor{darkgreen}{RGB}{40, 150, 40}
\setlist[itemize]{itemsep=0mm, topsep=2pt}

\AtBeginDocument{%
   \setlength\abovedisplayskip{3pt}
   \setlength\belowdisplayskip{3pt}}

\ShortHeadings{Improving Bayesian Optimization for Bipedal Robots}{Rai, Antonova, Meier, Atkeson}
\firstpageno{1}

\begin{document}

\title{Using Simulation to Improve Sample-Efficiency of Bayesian Optimization for Bipedal Robots}

\author{\name{Akshara Rai}$^{\pmb{\ast}}$
        \email arai@cs.cmu.edu\\
        \addr Robotics Institute, School of Computer Science\\
        Carnegie Mellon University, PA, USA
        \AND
        \name{Rika Antonova}$^{\pmb{\ast}}$
        \email antonova@kth.se\\
        \addr Robotics, Perception and Learning, CSC\\
        KTH Royal Institute of Technology, Stockholm, Sweden
        \AND
        \name Franziska Meier 
        \email 	franzi.meier@gmail.com\\
        \addr Paul G. Allen School of Computer Science \& Engineering \\ University of Washington, Seattle, WA, USA
        \AND
        \name Christopher G. Atkeson 
        \email cga@cs.cmu.edu\\
        \addr Robotics Institute, School of Computer Science\\
        Carnegie Mellon University, PA, USA\\
        {\scriptsize \normalfont $^{\pmb{\ast}}$These authors contributed equally.}
        }

\editor{}

\maketitle

\begin{abstract}
Learning for control can acquire controllers for novel robotic tasks, paving the path for autonomous agents. Such controllers can be expert-designed policies, which typically require tuning of parameters for each task scenario. In this context, Bayesian optimization (BO) has emerged as a promising approach for automatically tuning controllers. However, when performing BO on hardware for high-dimensional policies, sample-efficiency can be an issue. Here, we develop an approach that utilizes simulation to map the original parameter space into a domain-informed space. During BO, similarity between controllers is now calculated in this transformed space. Experiments on the ATRIAS robot hardware and another bipedal robot simulation show that our approach succeeds at sample-efficiently learning controllers for multiple robots. Another question arises: What if the simulation significantly differs from hardware? To answer this, we create increasingly approximate simulators and study the effect of increasing simulation-hardware mismatch on the performance of Bayesian optimization. We also compare our approach to other approaches from literature, and find it to be more reliable, especially in cases of high mismatch. Our experiments show that our approach succeeds across different controller types, bipedal robot models and simulator fidelity levels, making it applicable to a wide range of bipedal locomotion problems.
\end{abstract}


\input{introduction}
\input{background_bo}
\input{background_locomotion_ctrl}
\input{background_simulation}
\input{proposed_bo_with_informed_kernel}

\input{proposed_dog_transform}

\input{proposed_traj_nn}
\input{proposed_hwadjust}

\input{robot_and_controllers}

\input{simulator_versions}
\input{hw_experiments}
\input{biped7link_experiments}

\input{mismatch_experiments}
\input{conclusion}

\acks{
This research was supported in part by National Science Foundation grant IIS-1563807, the Max-Planck-Society, \& the Knut and Alice Wallenberg Foundation. Any opinions, findings, and conclusions or recommendations expressed in this material are those of the author(s) and do not necessarily reflect the views of the funding organizations.}

\appendix
\input{appendix_A}

\vskip 0.2in
\bibliography{references}

\end{document}

%% file: introduction.tex
\section{Introduction}
\label{sec:intro}
Machine learning can provide methods for learning controllers for robotic tasks. Yet, even with recent advances in this field, the problem of automatically designing and learning controllers for robots, especially bipedal robots, remains a difficult problem. Some of the core challenges of learning for control scenarios can be summarized as follows: It is expensive to do learning experiments that require a large number of samples with physical robots. Specifically, legged robots are not robust to falls and failures, and are time-consuming to work with and repair. Furthermore, commonly used cost functions for optimizing controllers are noisy to evaluate, non-convex and non-differentiable. In order to find learning approaches that can be used on real robots, it is thus important to keep these considerations in mind.

Deep reinforcement learning approaches can deal with noise, discontinuities and non-convexity of the objective, but they are not data-efficient. These approaches could take on the order of a million samples to learn locomotion controllers~\citep{peng2016terrain}, which would be infeasible on a real robot. For example, on the ATRIAS robot, $10,000$ samples would take $7$ days, in theory. But practically, the robot needs to be ``reset" between trials and repaired in case of damage. Using structured expert-designed policies can help minimize damage to the robot and make the search for successful controllers feasible. However, the problem is black-box, non-convex and discontinuous. This eliminates approaches like PI$^2$~\citep{theodorou2010generalized} which make assumptions about the dynamics of the system and PILCO~\citep{deisenroth2011pilco} which assumes a continuous cost landscape. Evolutionary approaches like CMA-ES~\citep{hansen2006cma} can still be prohibitively expensive, needing thousands of samples \citep{song2015neural}.

\begin{wrapfigure}{r}{0.31\textwidth}
\vspace{-0.1cm}
\includegraphics[width=0.3\textwidth]{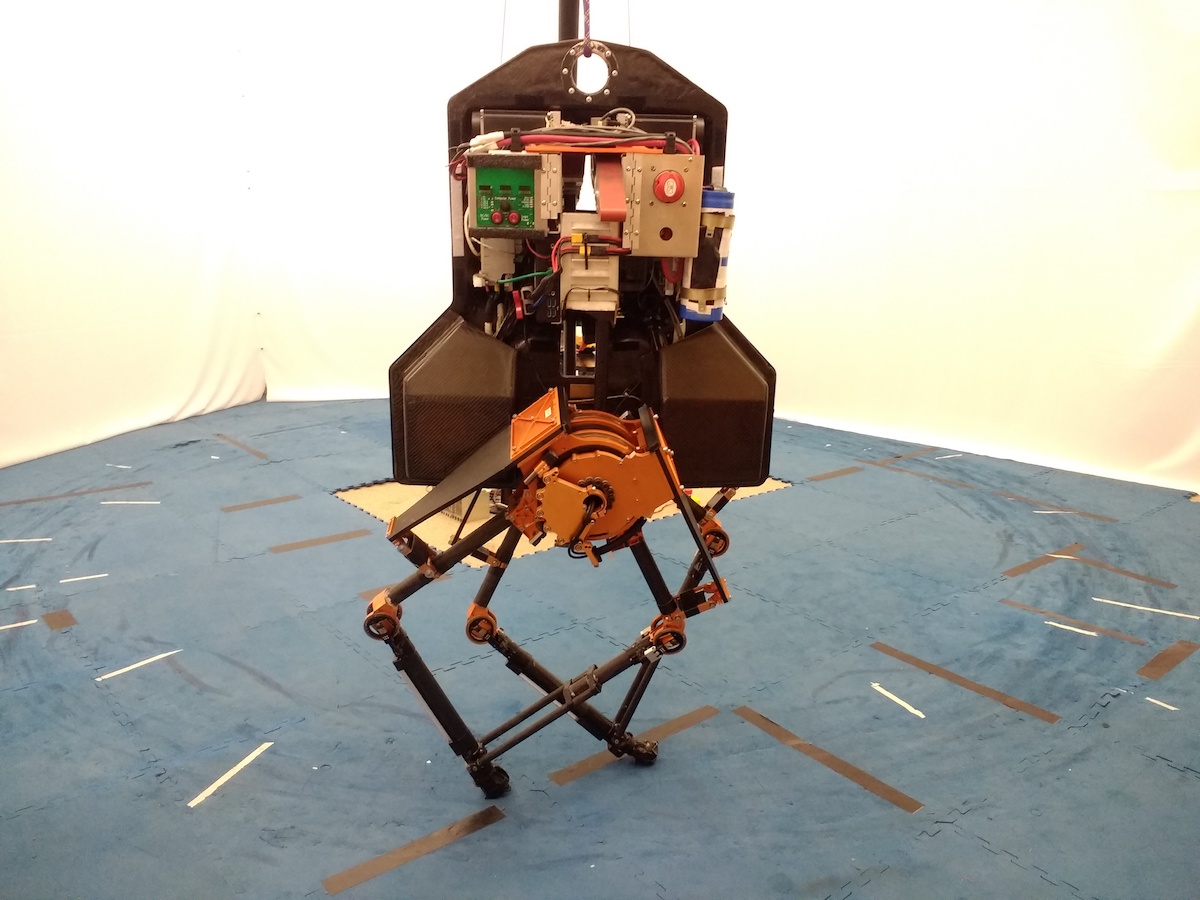}
\includegraphics[width=0.3\textwidth]{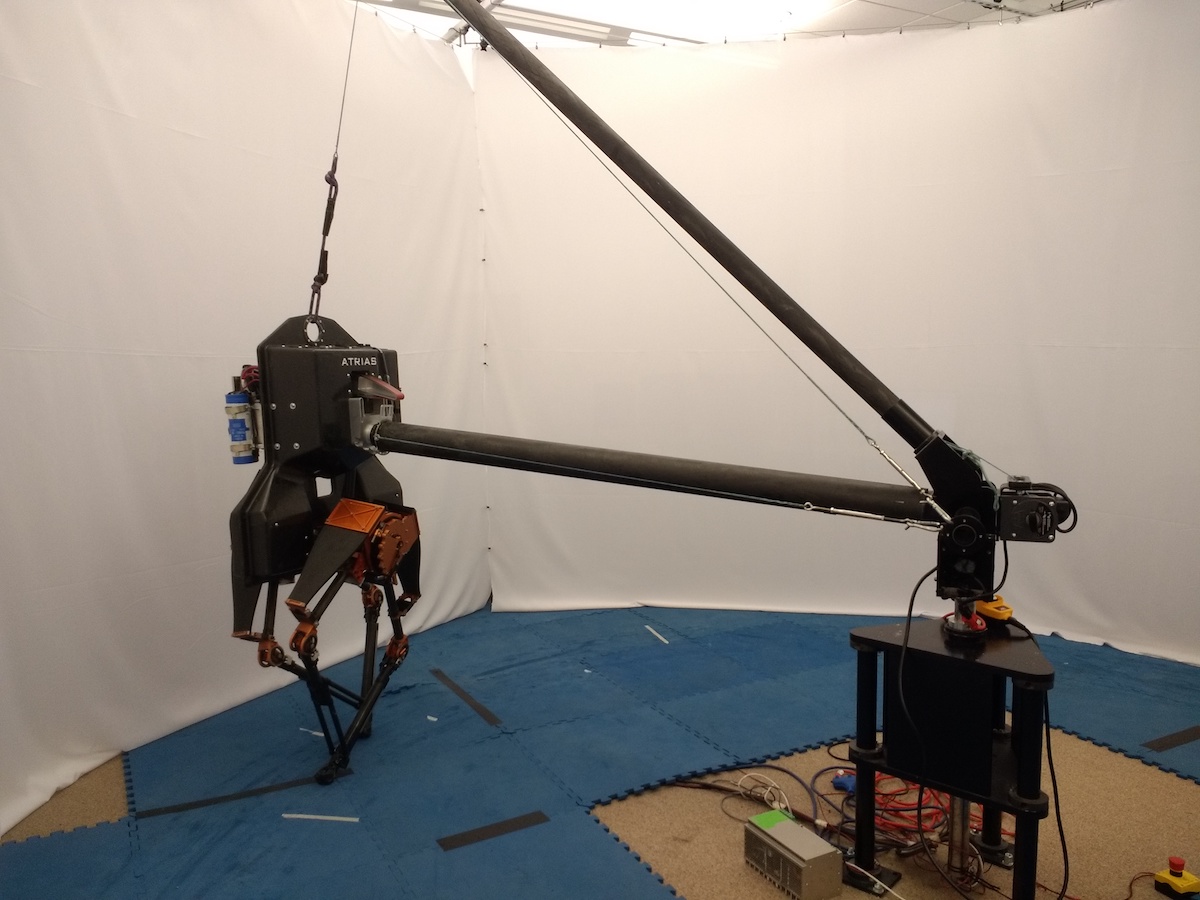}
\caption{\small{ATRIAS robot. }}
\label{fig:atrias}
\vspace{-0.3cm}
\end{wrapfigure}
In comparison, Bayesian optimization (BO) is a sample-efficient optimization technique that is robust to non-convexity, noise and even discontinuities. It has been recently used in a range of robotics problems, such as~\citet{calandra2016bayesian}, ~\citet{marco2017virtual}, ~\citet{cully2015robots}. However, sample-efficiency of conventional BO degrades in high dimensions, even for dimensionalities commonly encountered in locomotion controllers. Because of this, hardware-only optimization becomes intractable for  flexible controllers and complex robots. One way of addressing this issue is to utilize simulation to optimize controller parameters. However, simulation-only optimization is vulnerable to learning policies that exploit the simulation and perform well in simulation but poorly on the actual robot. This motivates the development of approaches that can incorporate simulation-based information into the learning method, then optimize with few samples on hardware. 

Towards this goal, our previous work in~\citet*{rai2016sample}, ~\citet{antonova2017deep}, ~\citet{rai2017bayesian} presents a framework that uses information from high-fidelity simulators to learn sample-efficiently on hardware. We use simulation to build informed feature transforms that are used to measure similarity during BO. Thus, the similarity between controller parameters, during optimization on hardware, is informed by how they perform in simulation. With this, it becomes possible to quickly infer which regions of the input space are likely to perform well on hardware. This method has been tested on the ATRIAS biped robot (Figure~\ref{fig:atrias}) and shows considerable improvement in sample-efficiency over traditional BO. 

In this article, we present in-depth explanations and empirical analysis of our previous work. Furthermore, for the first time, we present a procedure for systematically evaluating robustness of such approaches to simulation-hardware mismatch. We extend our previous work incorporating mismatch estimates~\citep{rai2017bayesian} to this setting. We also conduct extensive comparisons with competitive baselines from related work, such as~\citep{cully2015robots}. 

The rest of this article is organized as follows: Section~\ref{sec:background} provides background for BO, then gives an overview of related work on optimizing locomotion controllers.  Section~\ref{subsec:bo_with_informed_kernel_transform} describes the idea of incorporating simulation-based transforms into BO; Section~\ref{subsec:proposed_hwadjust} explains how we handle simulation-hardware mismatch.
Sections~\ref{subsec:atrias}-\ref{subsec:VNMC_cont} describe the robot and controllers we use for our experiments; Section~\ref{subsec:simulator_versions} explains the motivation and construction of simulators with various levels of fidelity. Section~\ref{sec:hw_experiments} gives a summary of hardware experiments conducted on the ATRIAS robot. Section~\ref{sec:biped_7link_experiments} shows generalization to a different robot model in simulation. Section~\ref{subsec:mismatch_experiments} shows empirical analysis of the impact of simulator fidelity on the performance of the proposed algorithms and alternative approaches.

%% file: background_bo.tex
\section{Background and Related Work}
\label{sec:background}
This section gives a brief overview of Bayesian optimization (BO), the state-of-the-art research on optimizing locomotion controllers, and utilizing simulation information in BO. 
\subsection{Background on Bayesian Optimization}
\label{sec:background_bo}

Bayesian optimization (BO) is a framework for online, black-box, gradient-free global search (\cite{BOtutorial2016} and~\cite{BOtutorial2010} provide a comprehensive introduction). The problem of optimizing controllers can be interpreted as finding controller parameters $\pmb{x}^*$ that optimize some cost function $f(\pmb{x})$. Here $\pmb{x}$ contains parameters of a pre-structured policy; the cost $f(\pmb{x})$ is a function of the trajectory induced by controller parameters $\pmb{x}$. For brevity, we will refer to `controller parameters $\pmb{x}$' as `controller $\pmb{x}$'. We use BO to find controller $\pmb{x}^*$, such that:
$\displaystyle f(\pmb{x}^*) = \min_{\pmb{x}} f(\pmb{x})$.

BO is initialized with a prior that expresses the \textit{a priori} uncertainty over the value of $f(\pmb{x})$ for each $\pmb{x}$ in the domain. Then, at each step of optimization, based on data seen so far, BO optimizes an auxiliary function (called \textit{acquisition function}) to select the next $\pmb{x}$ to evaluate. The acquisition function balances exploration vs exploitation. It selects points for which the posterior estimate of the objective $f$ is promising, taking into account both mean and covariance of the posterior. A widely used representation for the cost function $f$ is a Gaussian process (GP):
\vspace{-10px}
\begin{align*}
f(\pmb{x}) \sim \mathcal{GP}(\mu(\pmb{x}), k(\pmb{x}_i, \pmb{x}_j))
\end{align*}
The prior mean function $\mu(\cdot)$ is set to $0$ when no domain-specific knowledge is provided, or can be informative in the presence of information. The kernel function $k(\cdot, \cdot)$ encodes similarity between inputs. If $k(\pmb{x}_i, \pmb{x}_j)$ is large for inputs $\pmb{x}_i, \pmb{x}_j$, then $f(\pmb{x}_i$) strongly influences $f(\pmb{x}_j)$.
One of the most widely used kernel functions is the Squared Exponential (SE):
\begin{equation*}
k_{SE}(\pmb{x}_i, \pmb{x}_j) = \sigma_k^2 \exp\big(- \tfrac{1}{2} (\pmb{x}_i - \pmb{x}_j)^T \diag(\pmb{\ell})^{\!-\!2} (\pmb{x}_i - \pmb{x}_j) \big),
\end{equation*}
where $\sigma_k^2, \ \pmb{\ell}$ are signal variance and a vector of length scales respectively. $\sigma_k^2, \ \pmb{\ell}$ are referred to as `hyperparameters' in the literature. 

%% file: background_locomotion_ctrl.tex
\subsection{Optimizing Locomotion Controllers}
\label{subsec:background_locomotion_ctrl}

Parametric locomotion controllers can be represented as $\pmb{u} = \pi_{\pmb{x}}(\pmb{s})$, where $\pi$ is a policy structure that depends on parameters $\pmb{x}$. For example, $\pi$ can be parameterized by feedback gains on the center of mass (CoM), reference joint trajectories, etc. Vector $\pmb{s}$ is the state of the robot, such as joint angles and velocities; used in closed-loop controllers. Vector $\pmb{u}$ represents the desired control action, for example: torques, angular velocities or positions for each joint on the robot. The sequence of control actions yields a sequence of state transitions, which form the overall `trajectory' $[\pmb{s}_0, \pmb{u}_1, \pmb{s}_1, \pmb{u}_2, \pmb{s}_2, ...]$. This trajectory is used in the cost function to judge the quality of the controller $\pmb{x}$.
In our work, we use structured controllers designed by experts. State of the art research on walking robots featuring such controllers includes
\citet{feng2015optimization}, \citet{kuindersma2016optimization}. The overall optimization then includes manually tuning the parameters $\pmb{x}$. An alternative to manual tuning is to use evolutionary approaches, like \mbox{CMA-ES}, as in \citet{song2015neural}. However, these require a large number of samples and can usually be conducted only in simulation. Optimization in simulation can produce controllers that perform well in simulation, but not on hardware. 
In comparison, BO is a sample-efficient technique which has become popular for direct optimization on hardware. Recent successes include manipulation~\citep{englert2016combined} and locomotion~\citep{calandra2016bayesian}.

BO for locomotion has been previously explored for several types of mobile robots. These include: snake robots \citep{tesch}, AIBO quadrupeds \citep{lizotte2007automatic}, and hexapods \citep{cully2015robots}. \cite{tesch} optimize a 3-dimensional controller for a snake robot in 10-40 trials (for speeds up to $0.13m/s$).
\cite{lizotte2007automatic} use BO to optimize gait parameters for a AIBO robot in 100-150 trials. 
\cite{cully2015robots} learn 36 controller parameters for a hexapod. Even with hardware damage, they can obtain successful controllers for speeds up to $0.4m/s$ in 12-15 trials.

Hexapods, quadrupeds and snakes spend a large portion of their gaits being statically stable. In contrast, bipedal walking can be highly dynamic, especially for point-feet robots like ATRIAS. ATRIAS can only be statically stable in double-stance, and like most bipeds, spends a significant time of its gait being ``unstable", or dynamically stable. In our experiments on hardware, ATRIAS goes up to speeds of $1m/s$. All of this leads to a challenging optimization setting and discontinuous cost function landscape.
\cite{calandra2016bayesian} use BO for optimizing gaits of a dynamic biped on a boom, needing 30-40 samples for finding walking gaits for a 4-dimensional controller. While this is promising, optimizing a higher-dimensional controller needed for complex robots 
would be even more challenging. If significant number of samples lead to unstable gaits and falls, they could damage the robot. Hence, it is important to develop methods that can learn complex controllers fast, without damaging the robot. 

%% file: background_simulation.tex
\subsection{Incorporating Simulation Information into Bayesian Optimization}
\label{subsec:background_simulation}

The idea of using simulation to speed up BO on hardware has been explored before.
\cite{marco2017virtual} use simulation as a second source of noisy data. Information from simulation can also be added as a prior to the GP used in BO, such as in \cite{cully2015robots}. 
While these methods can be successful, one needs to carefully tune the influence of simulation points over hardware points, especially when simulation is significantly different from hardware.

Recently, several approaches proposed incorporating Neural Networks (NNs) into the Gaussian process (GP) kernels (\citet{wilson2016deep}, \citet{calandra2016manifold}). The strength of these approaches is that they can jointly update the GP and the NN. ~\citet{calandra2016manifold} demonstrated how this added flexibility can handle discontinuities in the cost function landscape. However, these approaches do not directly address the problem of incorporating a large amount of data from simulation in hardware BO experiments. 

\citet{wilson2014using} explored enhancing GP kernel with trajectories. Their Behavior Based Kernel (BBK) computes an estimate of a symmetric variant of the KL divergence between trajectories induced by two controllers, and uses this as a distance metric in the kernel. However, getting an estimate would require samples for each controller $\pmb{x}_i, \pmb{x}_j$ whenever $k(\pmb{x}_i, \pmb{x}_j)$ is needed. This can be impractical, as it involves an evaluation of every controller considered. The authors suggest combining BBK with a model-based approach to overcome this issue by learning a model. But building a reliable model might be an expensive process in itself. 

\citet{cully2015robots} utilize simulation by defining a behavior metric and collecting best performing points in simulation. This behavior metric then guides BO to quickly find controllers on hardware, and can even compensate for damage to the robot. The search on hardware is conducted in behavior space, and limited to pre-selected ``successful" points from simulation. This helps make their search faster and safer on hardware. However, if an optimal point was not pre-selected, BO cannot sample it during optimization.

In our work we develop two alternative strategies that utilize trajectories from simulation to build feature transforms that can be incorporated in the GP kernel used for BO. Our approaches incorporate trajectory/behavior information, but ensure that $k(\pmb{x}_i, \pmb{x}_j)$ is also computed efficiently during BO. They bias the search towards regions that look promising, but are able to `recover' and search in other parts of the space if simulation-hardware mismatch becomes apparent. 

%% file: proposed_bo_with_informed_kernel.tex
\section{Proposed Approach: Bayesian Optimization with Informed Kernels}
\label{sec:bo_with_informed_kernel}

\begin{figure}[t]
    \centering
    \includegraphics[width=0.8\textwidth]{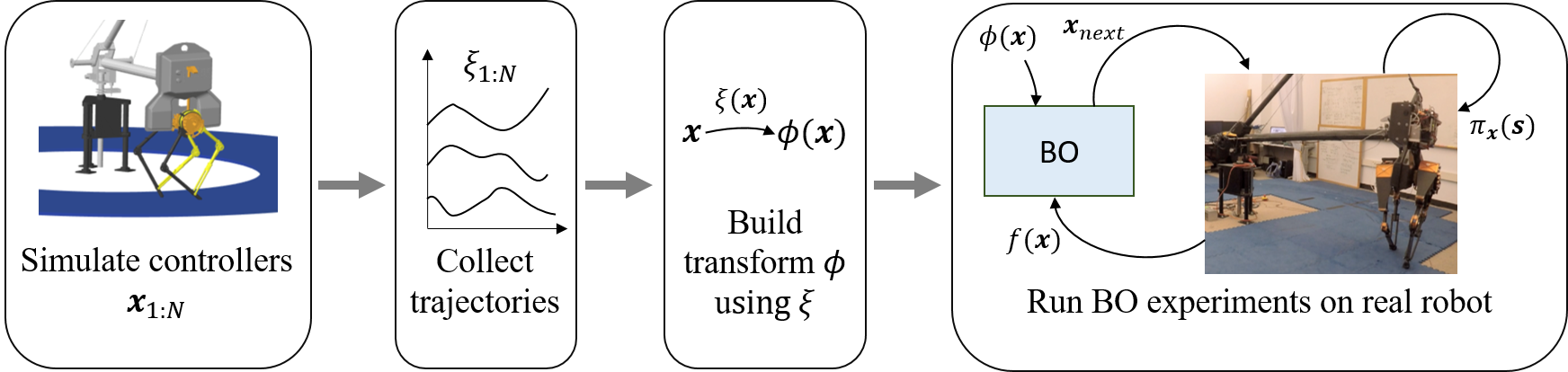}
    \caption{Overview of our proposed approach. Here, $\pi_{\pmb{x}}(\pmb{s})$ is the policy (Section~\ref{subsec:background_locomotion_ctrl});  $\pmb{x}$ is a vector of controller parameters; $\pmb{s}$ is the state of the robot; $\xi(\pmb{x})$ is a trajectory observed in simulation for $\pmb{x}$; $\phi(\cdot)$ is the transform built using $\xi(\pmb{x})$. $f(\pmb{x})$ is the cost of $\pmb{x}$ evaluated on hardware. BO uses $\phi(\pmb{x})$ and evaluated costs $f(\pmb{x})$ to propose next promising controller $x_{next}$.}
    \label{fig:approach}
    \vspace{-0.5cm}
\end{figure}

In this section, we offer in-depth explanation of approaches from our work in~\citet*{rai2016sample}, ~\citet{antonova2017deep}, and~\citet{rai2017bayesian}. This work proposes incorporating domain knowledge into BO with the help of simulation. We evaluate locomotion controllers in simulation, and collect their induced trajectories, which are then used to build an informed transform. This can be achieved by using a domain-specific feature transform (Section~\ref{subsec:proposed_dog_transform}) or by learning to reconstruct short trajectory summaries (Section~\ref{subsec:proposed_traj_nn}). This feature trasform is used to construct an informed distance metric for BO, and helps BO discover promising regions faster. An overview can be found in Figure \ref{fig:approach}. In Section~\ref{subsec:proposed_hwadjust} we discuss how to incorporate simulation-hardware mismatch in to the transform, ensuring that BO can benefit from inaccurate simulations as well.

\subsection{Constructing Flexible Kernels using Simulation-based Transforms}
\label{subsec:bo_with_informed_kernel_transform}
High dimensional problems with discontinuous cost functions are very common with legged robots, where slight changes to some parameters can make the robot unstable. Both of these factors can adversely affect BO's performance, but informed feature transforms can help BO sample high-performing controllers even in such scenarios.

In this section, we demonstrate how to construct such transforms $\phi(\pmb{x})$ utilizing simulations for a given controller $\pmb{x}$. We then use $\phi$ to create an informed kernel $k_{\phi}(\pmb{x}_i, \pmb{x}_j)$ for BO on hardware:

\vspace{-2px}
\begin{align}
\begin{split}
    \pmb{t}_{ij} \!=\! \phi(\pmb{x}_i) \!-\! \phi(\pmb{x}_j) \\
    k_{\phi}(\pmb{x}_i, \pmb{x}_j) = \sigma_k^2 \exp\Big(- \tfrac{1}{2} \pmb{t}_{ij}^T \diag(\pmb{\ell})^{\!-\!2} \pmb{t}_{ij} \Big)
\end{split}
\end{align}
Note that the functional form above is same as that of Squared Exponential kernel, if considered from the point of view of the transformed space, with $\phi(\pmb{x})$ as input. While this kernel is stationary as a function of $\phi$, it is non-stationary in $\pmb{x}$. $\phi$ can bring closer related parts of the space that would be otherwise far apart in the original space. BO can then operate in the space of $k_{\phi}$, which is `informed' by simulation.

%% file: proposed_dog_transform.tex
\subsubsection{The Determinants of Gait Transform}
\label{subsec:proposed_dog_transform}

We propose a feature transform for bipedal locomotion derived from physiological features of human walking called Determinants of Gaits (DoG) \citep{inman1953major}. $\phi_{DoG}$ was originally developed for human-like robots and controllers \citep*{rai2016sample}, and then generalized to be applicable to a wider range of bipedal locomotion controllers and robot morphologies \citep{rai2017bayesian}. It is based on the features in Table~\ref{tbl:dog_features}.

\begin{table*}[h] 
\centering
\small{
\begin{tabular}{ @{}l@{}  p{12.5cm} }
\toprule
\multirow{ 3}{*}{
\centering
\includegraphics[width=0.15\textwidth]{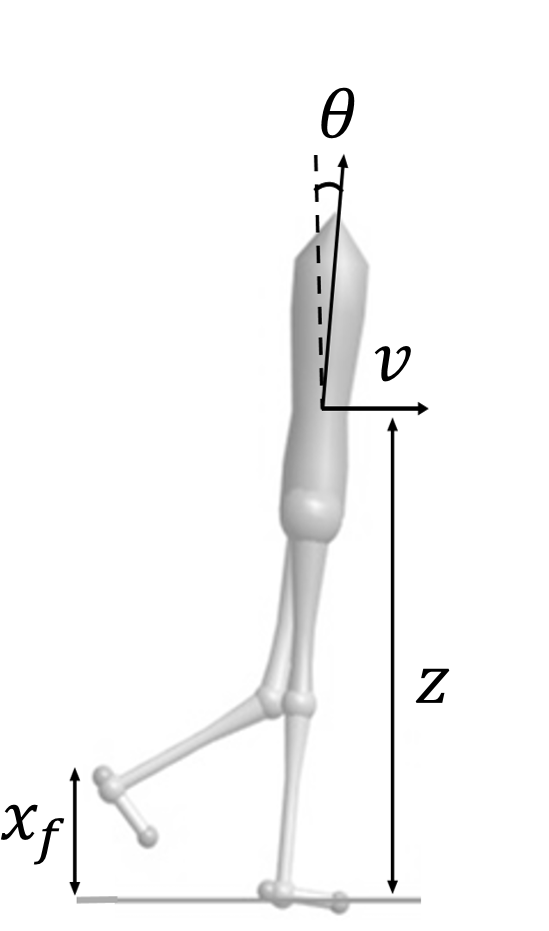}
} & $M_1$ (Swing leg retraction) -- If the maximum ground clearance of the swing foot $x_f$ is more than a threshold, $M_1\!=\!1$ (0 otherwise); ensures swing leg retraction. \\
& \\
 &  $M_2$ (Center of mass height) -- If CoM height $z$ stays about the same at the start and end of a step, $M_2\!=\!1$ (0 otherwise); checks that the robot is not falling. \\
 & \\
 & $M_3$ (Trunk lean) -- If the average trunk lean $\theta$ is the same at the start and end of a step, $M_3\!=\!1$ (0 otherwise); ensures that the trunk is not changing orientation.\\
 & \\
 & $M_4$ (Average walking speed) -- Average forward speed $v$ of a controller per step, $M_4\!=\!v_{avg}$; $M_4$ helps distinguish controllers that perform similar on $M_{1-3}$. \\
 \bottomrule
\end{tabular}
} % end small
\caption{Illustration of the features used to construct DoG transform.}
\label{tbl:dog_features}
\vspace{-0.25cm}
\end{table*}

$\phi_{DoG}$ combines features $M_{1-4}$ per step and scales them by the normalized simulation time to obtain the DoG score of controller $\pmb{x}$:
\vspace{-5px}
\begin{align}
\label{eq:dog}
score_{DoG} = \frac{t_{sim}}{t_{max}} \cdot \sum_{s=1}^{N} \sum_{k=1}^4 M_{k s}
\end{align}
Here $N$ is the number of steps taken in simulation, $t_{sim}$ is time at which simulation terminated (possibly due to a fall), $t_{max}$ is total time allotted for simulation. Since larger number of steps lead to higher DoG, some controllers that chatter (step very fast before falling) could get misleadingly high scores; we scale the scores by $\frac{t_{sim}}{t_{max}}$ to prevent that. 
$\phi_{DoG}(\pmb{x})$ for controller parameters $\pmb{x}$ now becomes the computed $score_{DoG}$ of the resulting trajectories when $\pmb{x}$ is simulated. $\phi_{DoG}$ essentially aids in (soft) clustering of controllers based on their behaviour in simulation. High scoring controllers are more likely to walk than low scoring ones. Since $M_{1-4}$ are based on intuitive gait features, they are more likely to transfer between simulation and hardware, as compared to direct cost. The thresholds in $M_{1-3}$ are chosen according to values observed in nominal human walking from \cite{winter1987emg}.

%% file: proposed_traj_nn.tex
\subsubsection{Learning Feature Transform with a Neural Network}
\label{subsec:proposed_traj_nn}

While domain-specific feature transforms can be extremely useful and robust, they might be difficult to generate when a domain expert is not present. This motivates directly learning such feature transforms from trajectory data. In this section we describe our approach to train neural networks to reconstruct trajectory summaries~\citep{antonova2017deep} that achieves this goal of minimizing expert involvement.

Trajectory summaries are a convenient choice for reparametrizing controllers into an easy to optimize space. For example, controllers that fall would automatically be far away from controllers that walk. If these trajectories can be extracted from a high-fidelity simulator, we would not have to evaluate each controller on hardware. However, conventional implementations of BO evaluate the kernel function for a large number of points per iteration, requiring thousands of simulations each iteration. To avoid this, a Neural Network (NN) can be trained to reconstruct trajectory summaries from a large set of pre-sampled data points. 
NN provides flexible interpolation, as well as fast evaluation (controller $\rightarrow$ trajectory summary). Furthermore, trajectories are agnostic to the specific cost used during BO. Thus the data collection can be done offline, and there is no need to re-run simulations in case the definition of the cost is modified.

We use the term `trajectory' in a general sense, referring to several sensory states recorded during a simulation. To create trajectory summaries for the case of locomotion, we include measurements of: walking time (time before falling), energy used during walking, position of the center of mass and angle of the torso. With this, we construct a dataset for NN to fit: a Sobol grid of controller parameters ($\pmb{x}_{1:N}$, $N\!\approx\!0.5$ million) along with trajectory summaries $\xi_{\pmb{x}_i}$ %$\pmb{traj}_{\pmb{x}_i}$ 
from simulation. NN is trained using mean squared loss:

\ \ \ \ NN input: $\pmb{x}$ -- a set of controller parameters

\ \ \ \ NN output: 
$\phi_{\textit{trajNN}}(\pmb{x}) = \hat{\xi}_{\pmb{x}}$ -- reconstructed trajectory summary 

\ \ \ \ NN loss: $\tfrac{1}{2} \sum_{i=1}^N || \hat{\xi}_{\pmb{x}_i} - \xi_{\pmb{x}_i}||^2$

\noindent The outputs $\phi_{\textit{trajNN}}(\pmb{x})$ are then used in the kernel for BO:
\begin{equation}
k_{\textit{trajNN}}(\pmb{x}_i, \pmb{x}_j) = \sigma_k^2 \exp\big(\!-\!\tfrac{1}{2} \pmb{t}_{ij}^T \diag( \pmb{\ell})^{\!-\!2} \pmb{t}_{ij}\big), \quad \quad \pmb{t}_{ij} = \phi_{\textit{trajNN}}(\pmb{x}_i) - \phi_{\textit{trajNN}}(\pmb{x}_j) 
\end{equation}
We did not carefully select the sensory traces used in the trajectory summaries. Instead, we used the most obvious states, aiming for an approach that could be easily adapted to other domains. To apply this approach to a new setting, one could simply include information that is customarily tracked, or used in costs. For example, for a manipulator, the coordinates of the end effector(s) could be recorded at relevant points. Force-torque measurements could be included, if available.

%% file: proposed_hwadjust.tex
\subsection{Kernel Adjustment for Handling Simulation-Hardware Mismatch}
\label{subsec:proposed_hwadjust}

Approaches described in previous sections could provide improvement for BO when a high-fidelity simulator is used in kernel construction. In~\citet{rai2017bayesian} we presented promising results of experimental evaluation on hardware. However, it is unclear how the performance changes when simulation-hardware mismatch becomes apparent. 

In~\citet{rai2017bayesian}, we also proposed a way to incorporate information about simulation-hardware mismatch into the kernel from the samples evaluated so far. We augment the simulation-based kernel to include this additional information about mismatch, by expanding the original kernel by an extra dimension that contains the predicted mismatch for each controller $\pmb{x}$.
 
A separate Gaussian process is used to model the mismatch experienced on hardware, starting from an initial prior mismatch of 0: $g(\pmb{x})\!\sim\!\mathcal{GP}(0, k_{SE}(\pmb{x}_i,\pmb{x}_j))$.
For any evaluated controller $\pmb{x}_i$, we can compute the difference between $\phi(\pmb{x}_i)$ in simulation and on hardware: $d_{\pmb{x}_i}\!=\!\phi^{sim}(\pmb{x}_i) \!-\! \phi^{hw}(\pmb{x}_i)$. 
We can now use mismatch data $\{ d_{\pmb{x}_i} | i=1...n \}$ to construct a model for the expected mismatch: $\bar{g}(\pmb{x})$. In the case of using a GP-based model, $\bar{g}(\cdot)$ would denote the posterior mean. With this, we can predict simulation-hardware mismatch in the original space of controller parameters for unevaluated controllers. Combining this with kernel $k_{\phi}$ we obtain an adjusted kernel:
\begin{align}
\label{eq:k_adj}
\begin{split}
    \pmb{\phi}^{adj}_{\pmb{x}} = \begin{bmatrix} \phi^{sim}(\pmb{x}) \\ \bar{g}(\pmb{x}) \end{bmatrix}, \quad \quad \quad \quad \quad \quad 
    \pmb{t}_{ij}^{adj} \!=\! \pmb{\phi}^{adj}_{\pmb{x}_i} \!-\! \pmb{\phi}^{adj}_{\pmb{x}_j} \\
    k_{\phi_{adj}}(\pmb{x}_i, \pmb{x}_j) = \sigma_k^2 \exp\Big(- \tfrac{1}{2} (\pmb{t}_{ij}^{adj})^T \diag\Big(\begin{bmatrix}\pmb{\ell_1} \\ \pmb{\ell_2} \end{bmatrix}\Big)^{\!-\!2} \pmb{t}_{ij}^{adj} \Big) \\
\end{split}
\end{align}
The similarity between points $\pmb{x}_i, \pmb{x}_j$ is now dictated by two components: representation in $\phi$ space and expected mismatch. This construction has an intuitive explanation: Suppose controller $\pmb{x_i}$ results in walking when simulated, but falls during hardware evaluation. $k_{\phi_{adj}}$ would register a high mismatch for $\pmb{x}_i$. Controllers would be deemed similar to $\pmb{x}_i$ only if they have both similar simulation-based $\phi(\cdot)$ and similar estimated mismatch. Points with similar simulation-based $\phi(\cdot)$ and low predicted mismatch would still be `far away' from the failed $\pmb{x}_i$. This would help BO sample points that still have high chances of walking in simulation, but are in a different region of the original parameter space. In the next section, we present a more mathematically rigorous interpretation for $k_{\phi_{adj}}$.

\subsubsection{Interpretation of Kernel with Mismatch Modeling}

Let us consider a controller $\pmb{x}_i$ evaluated on hardware. The difference between simulation-based and hardware-based feature transform for $\pmb{x}_i$ is $d_{\pmb{x}_i} = \phi^{sim}(\pmb{x}_i) - \phi^{hw}(\pmb{x}_i)$. The `true' hardware feature transform for $\pmb{x_i}$ is $\phi^{hw}(\pmb{x_i}) = \phi^{sim}(\pmb{x_i}) - d_{\pmb{x_i}}$. After $n$ evaluations on hardware, $\{ d_{\pmb{x}_i} | i=1...n \}$ can serve as data for modeling simulation-hardware mismatch. In principle, any data-efficient model $g(\cdot)$ can be used, such as GP (a multi-output GP in case $\phi(\cdot)$ $>1D$). With this, we can obtain an adjusted transform: $\hat{\phi}^{hw}(\pmb{x}) = \phi^{sim}(\pmb{x}) - \bar{g}(\pmb{x})$, where $\bar{g}(\cdot)$ is the output of the model fitted using $\{d_{\pmb{x}_i} | i=1,...n \}$.

Suppose $\pmb{x}_{new}$ has not been evaluated on hardware. We can use $\hat{\phi}^{hw}(\pmb{x}_{new}) = \phi^{sim}(\pmb{x}_{new}) - \bar{g}(\pmb{x}_{new})$ as the adjusted estimate of what the output of $\phi$ should be, taking into account what we have learned so far about simulation-hardware mismatch.

Let's construct kernel $k^{v_2}_{\phi_{adj}}(\pmb{x}_i, \pmb{x}_j)$ that uses these hardware-adjusted estimates directly:
\begin{equation*}
\begin{split}
\pmb{q}_{ij}^{adj} &= \hat{\phi}^{hw}(\pmb{x}_{i}) - \hat{\phi}^{hw}(\pmb{x}_{j}) \\
&= (\phi^{sim}(\pmb{x}_{i}) - \bar{g}(\pmb{x}_i)) - (\phi^{sim}(\pmb{x}_{j}) - \bar{g}(\pmb{x}_j) ) \\
&= \underbrace{(\phi^{sim}(\pmb{x}_{i}) - \phi^{sim}(\pmb{x}_{j}))}_{\pmb{v}_{\phi}} + \underbrace{(\bar{g}(\pmb{x}_j) - \bar{g}(\pmb{x}_i) )}_{\pmb{v}_{g}}
\end{split}
\end{equation*}

\begin{equation*}
\begin{split}
k^{v_2}_{\phi_{adj}}(\pmb{x}_i, \pmb{x}_j) &= \sigma^2_{k_{v_0}} \exp\Big(- \tfrac{1}{2} (\pmb{q}_{ij}^{adj})^T \diag(\pmb{\ell})^{\!-\!2} \pmb{q}_{ij}^{adj} \Big) \\
& = \sigma^2 \exp\Big(- \tfrac{1}{2} \big[ (\pmb{v}_{\phi} + \pmb{v}_{g})^T \diag(\pmb{\ell})^{\!-\!2} (\pmb{v}_{\phi} + \pmb{v}_{g}) \big] \Big) \\
& = \sigma^2
\exp\Big(- \tfrac{1}{2} \big[ \pmb{v}_{\phi}^T \diag(\pmb{\ell})^{\!-\!2} \pmb{v}_{\phi} + \pmb{v}_{g}^T\diag(\pmb{\ell})^{\!-\!2} \pmb{v}_{g} + prod_{ij} \big] \Big)\\
&\quad \quad \quad \quad \quad \text{where } prod_{ij} = 2 \pmb{v}_{\phi}^T\diag(\pmb{\ell})^{\!-\!2} \pmb{v}_{g}
\end{split}
\end{equation*}

Using $\exp(a+b+c)=\exp(c) \cdot \exp(a+b)$,  we have:
\begin{align*}
k^{v_2}_{\phi_{adj}}(\pmb{x}_i, \pmb{x}_j) &= \sigma^2 \exp(-prod_{ij}) \exp\Big(- \tfrac{1}{2} \big[ \pmb{v}_{\phi}^T \diag(\pmb{\ell})^{\!-\!2} \pmb{v}_{\phi} + \pmb{v}_{g}^T\diag(\pmb{\ell})^{\!-\!2} \pmb{v}_{g} \big] \Big)
\end{align*}

If we now observe that $\pmb{v}_{g}^T\diag(\pmb{\ell})^{\!-\!2} \pmb{v}_{g} = (-\pmb{v}_{g})^T\diag(\pmb{\ell})^{\!-\!2} (-\pmb{v}_{g})$ we get:
\begin{align*}
k^{v_2}_{\phi_{adj}}(\pmb{x}_i, \pmb{x}_j) 
&= \sigma^2 \exp(-prod_{ij}) \exp\Big(- \tfrac{1}{2} (\pmb{t}_{ij}^{adj})^T \diag\Big(\begin{bmatrix}\pmb{\ell} \\ \pmb{\ell} \end{bmatrix} \Big)^{\!-\!2} \pmb{t}_{ij}^{adj} \Big) \\
\pmb{t}_{ij}^{adj} &= 
\begin{bmatrix}
\phi^{sim}(\pmb{x}_{i}) - \phi^{sim}(\pmb{x}_{j}) \\
\bar{g}(\pmb{x}_i) - \bar{g}(\pmb{x}_j)
\end{bmatrix} \text{ (from Equation~\ref{eq:k_adj})}
\end{align*}

Compare this to $k_{\phi_{adj}}$ from Equation~\ref{eq:k_adj}:
\begin{equation}
k_{\phi_{adj}}(\pmb{x}_i, \pmb{x}_j) = \sigma_k^2 \exp\Big(- \tfrac{1}{2} (\pmb{t}_{ij}^{adj})^T \diag\Big(\begin{bmatrix}\pmb{\ell_1} \\ \pmb{\ell_2} \end{bmatrix}\Big)^{\!-\!2} \pmb{t}_{ij}^{adj} \Big)
\end{equation}
Now we see that $k^{v_2}_{\phi_{adj}}$ and $k_{\phi_{adj}}$ have a similar form. Hyperparameters $\pmb{\ell_1}, \pmb{\ell_2}$ provide flexibility in $k_{\phi_{adj}}$ as compared to having only vector $\pmb{\ell}$ in $k^{v_2}_{\phi_{adj}}$. They can be adjusted manually or with Automatic Relevance Determination. 
For $k^{v_2}_{\phi_{adj}}$, the role of signal variance is captured by $\sigma^2 \exp(-prod_{ij})$. This makes the kernel nonstationary in the transformed $\phi$ space. 
Since $k_{\phi_{adj}}$ is already non-stationary in $\pmb{x}$, it is  unclear whether non-stationarity of $k^{v_2}_{\phi_{adj}}$ in the transformed $\phi$ space has any advantages.

The above discussion shows that $k_{\phi_{adj}}$ proposed in~\cite{rai2017bayesian} is motivated both intuitively and mathematically. It aims to use a transform that accounts for the hardware mismatch, without adding extra non-stationarity in the transformed space.

%----------------------------------------------------------------------------------------------

%% file: robot_and_controllers.tex
\section{Robots, Simulators and Controllers Used}
\label{sec:robot_and_controllers}

In this section we give a concise description of the robots, controllers and simulators used in experiments with BO for bipedal locomotion. We aim for our approach to be applicable to a wide range of bipedal robot morphologies and controllers, including state-of-the-art controllers \citep{feng2015optimization}. This ensures that our experimental results are relevant to current research for bipedal locomotion and are transferable to other systems. 

We work with two different types of controllers -- a reactively stepping controller and a human-inspired neuromuscular controller (NMC). The reactively stepping controller is model-based: it uses inverse-dynamics models of the robot to compute desired motor torques. In contrast, NMC is model-free: it computes desired torques using hand-designed policies, created with biped locomotion dynamics in mind. These controllers exemplify two different and widely used ways of controlling bipedal robots. In addition to this, we show results on two different robot morphologies -- a parallel bipedal robot ATRIAS, and a serial 7-link biped model. Our hardware experiments are conducted on ATRIAS; the 7-link biped is only used in simulation. Our success on both robots shows that the approaches developed in this paper are widely applicable to a range of bipedal robots and controllers.

\subsection{ATRIAS Robot}
\label{subsec:atrias}
Our hardware platform is an ATRIAS robot (Figure~\ref{fig:atrias}). ATRIAS is a parallel bipedal robot with most of its mass concentrated around the torso, weighing $\approx 64kg$. The legs are 4-segment linkages actuated by 2 Series Elastic Actuators (SEAs) in the sagittal plane and a DC motor in the lateral plane. Details can be found in \cite{hubicki2016atrias}.
In this work we focus on planar motion around a boom. ATRIAS is a highly dynamic system due to its point feet, with static stability only in double stance on the boom. 

\subsection{Planar 7-link Biped}
\label{sec:biped}

\begin{wrapfigure}{r}{0.091\textwidth}
\vspace{-14px}
    \centering
    \includegraphics[width = 0.09\textwidth]{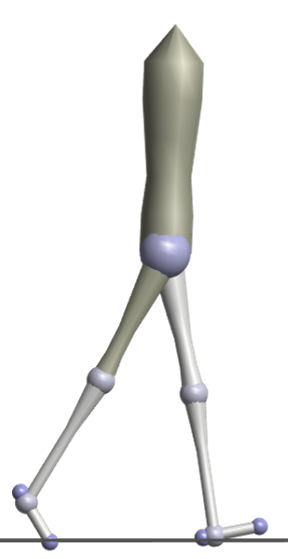}
    \caption{\small{\mbox{7-link} biped}}
    \label{fig:biped}
\vspace{-5px}
\end{wrapfigure}
The second robot used in our experiments is a 7-link biped (Figure~\ref{fig:biped}). It has a trunk and segmented legs with ankles. Unlike ATRIAS, this is a series robot with actuators in the hip, knees and ankles. The inertial properties of its links are similar to an average human \citep{winter1987emg}. 
The simulator code is modified from \cite{thatte2016}. The 7-link model is a canonical simulator for testing bipedal walking algorithms, for example in \citet{song2015neural}. It is a simplified two-dimensional simulator for a large range of humanoid robots, like Atlas \citep{feng2015optimization}. The purpose of using this simulator is to study the generalizability of our proposed approaches to systems different from ATRIAS.

%-------------------------------------------------------------------------------------
\subsection{Feedback Based Reactive Stepping Policy}
\label{subsec:raibert_cont}

We design a parametrized controller for controlling the CoM height, torso angle and the swing leg by commanding desired ground reaction forces and swing foot landing location. 
\begin{align*}
    \begin{split}
    F_x = K_{pt}(\theta_{des} \!- \theta) - K_{dt} \dot{\theta} \quad \quad
    F_z = K_{pz}(z_{des} \!- z) - K_{dz}\dot{z}\quad \quad 
    x_p = k(v-v_{tgt}) + C d + 0.5 v T
    \end{split}
\end{align*}

Here, $F_x$ is the desired horizontal ground reaction force (GRF), $K_{pt}$ and $K_{dt}$ are the proportional and derivative feedback gains on the torso angle $\theta$ and  velocity $\dot{\theta}$. $F_z$ is the desired vertical GRF, $K_{pz}$ and $K_{dz}$ are the proportional and derivative gains on the CoM height $z$ and vertical velocity $\dot{z}$. $z_{des}$ and $\theta_{des}$ are the desired CoM height and torso lean. $x_p$ is the desired foot landing location for the end of swing; $v$ is the horizontal CoM velocity, $k$ is the feedback gain that regulates $v$ towards the target velocity $v_{tgt}$.

$C$ is a constant and $d$ is the distance between the stance leg and the CoM; $T$ is the swing time.

The desired GRFs are sent to ATRIAS inverse dynamics model that generates desired motor torques $\tau_f, \tau_b$. Details can be found in \cite{rai2017bayesian}. 
 
This controller assumes no double-stance, and the swing leg takes off as soon as stance is detected. This leads to a highly dynamic gait, as the contact polygon for ATRIAS in single stance is a point, posing a challenging optimization problem. 

To investigate the effects of increasing dimensionality on our optimization, we construct two controllers with different number of free parameters:
\begin{itemize}
\item\textit{5-dimensional controller} : optimizing 5 parameters $[K_{pt}, K_{dt}, k, C, T]$ \\
($z_{des}$, $\theta_{des}$ and the feedback on $z$ are hand tuned)
\item \textit{9-dimensional controller} : optimizing all 9 parameters of the high-level policy\\
$[K_{pt}, K_{dt}, \theta_{des}, K_{pz}, K_{dz}, z_{des}, k, C, T]$
\end{itemize}

%-------------------------------------------------------------------------------------
\subsection{16-dimensional Neuromuscular Controller}
\label{sec:nmm_cont}

We use neuromuscular model policies, as introduced in \cite{geyer2010muscle}, as our controller for the 7-link planar human-like biped model. These policies use approximate models of muscle dynamics and human-inspired reflex pathways to generate joint torques, producing gaits that are similar to human walking.% 
 
Each leg is actuated by 7 muscles, which together produce torques about the hip, knee and ankle. Most of the muscle reflexes are length or force feedbacks on the muscle state aimed at generating a compliant leg, keeping knee from hyperextending and maintaining torso orientation in stance. The swing control has three main components -- target leg angle, leg clearance and hip control due to reaction torques. Together with the stance control, this leads to a total of 16 controller parameters, described in details in \cite{rai2016sample}. 

Though originally developed for explaining human neural control pathways, this controller has recently been applied to prosthetics and bipeds, for example~\citet{thatte2016} and~\citet{van2015experimental}. As demonstrated in \citet{song2015neural}, this controller is capable of generating a variety of locomotion behaviours for a humanoid model --  walking on rough ground, turning, running, and walking upstairs, making it a very versatile controller. This is a model-free controller as compared to the reactive-stepping controller, which was model-based.

%-------------------------------------------------------------------------------------
\subsection{50-dimensional Virtual Neuromuscular Controller}
\label{subsec:VNMC_cont}

Another model-free controller we use on ATRIAS is a modified version of~\cite{batts2015toward}.
VNMC maps a neuromuscular model, similar to the one described in Section \ref{sec:nmm_cont} to the ATRIAS robot's topology and emulates it to generate desired motor torques.
The robot's states are mapped to the states of a virtual 5-link bipedal robot. This virtual robot then used by VNMC to generate knee and hip torques which are then mapped back to the robot torques, in swing and stance.
We adapt VNMC by removing some biological components while preserving its basic functionalities.
First, the new VNMC directly uses joint angles and angular velocity data instead of estimating it from physiological sensory data, such as muscle fiber length and velocity. 
Second, most of the neural transmission delays are removed, except those utilized by the controller. The final version of the controller consists of 50 parameters including low-level control parameters, such as feedback gains, as well as high level parameters, such as desired step length and desired torso lean.  When optimized using CMA-ES, it can control ATRIAS to walk on rough terrains with height changes of $\pm$20 cm in planar simulation \citep{batts2015toward}.

%% file: simulator_versions.tex
\subsection{Simulators with Different Levels of Fidelity}
\label{subsec:simulator_versions}

To compare the performance of different methods that can be used to transfer information from simulation to hardware, we create a series of increasingly approximate simulators. These simulators emulate increasing mismatch between simulation and hardware and its effect on the information transfer.
In this setting, the high-fidelity ATRIAS simulator~\citep{martin2015robust}, which was used in all the previous simulation experiments becomes the \textit{simulated ``hardware"}. Next we make dynamics approximations to the original simulator, which are used commonly in simulators to decrease fidelity and increase simulation speed. For example, the complex dynamics of harmonic drives are approximated as a torque multiplication, and the boom is removed from the simulation, leading to a two-dimensional simulator. These approximate simulators now become the \textit{simulated ``simulators"}. As the approximations in these simulators are increased, we expect the performance of methods that utilize simulation for optimization on hardware to deteriorate.

\noindent The details of the approximate simulators are described in the two paragraphs below:

\noindent \textbf{1. Simulation with simplified gear dynamics} : The ATRIAS robot has geared DC motors attached to leaf springs on the legs. Their high gear ratio of 50 is achieved through a harmonic drive. In the original simulator, this drive is modelled using gear constraints in MATLAB SimScape Multibody simulation environment. These require significant computation time as the constraint equations have to be solved at every time instant, but lead to a very good match between the robot and simulation. We replace this model with a commonly used approximation for geared systems -- multiplying the rotor torque by the gear ratio. This reduces the simulation time to about a third of the original simulator, but leads to an approximate gear dynamics model.

\noindent \textbf{2. Simulation with no boom and simplified gear dynamics} : The ATRIAS robot walks on a boom in our hardware experiments. The boom leads to lateral torques on the robot, which have vertical and horizontal force components that need to be considered in a realistic simulation of the robot. In our second approximation, we remove the boom from the original simulator and constraint the motion of the robot to a 2-dimensional plane, making a truly two-dimensional simulation of ATRIAS. This is a common approximation for two-dimensional robots. Since this approximation has both simplified gear dynamics and no boom, it is further from the original simulator than the first approximation.

The advantage of such an arrangement is that we can extensively test the effect of un-modelled and wrongly modelled dynamics on information transfer between simulation and hardware. Even in our high-fidelity original simulator, there are several un-modelled components of the actual hardware. For example, the non-rigidness of the robot parts, misaligned motors and relative play between joints. In our experiments, we find that the 50-dimensional VNMC is a sensitive controller, with little hope of directly transferring from simulation to hardware. Anticipating this, we can now test several methods of compensating for this mismatch using our increasingly approximate simulators. In the future, we would like to take this approximations further and study when there is useful information even in over-simplified simulations of legged systems.

%% file: hw_experiments.tex
\section{Experiments}
\label{sec:hw_experiments}
We will now present our experiments on optimizing controllers that are 5, 9, 16 and 50 dimensional. 
We split our experiments into three categories: hardware experiments on the ATRIAS robot, simulation experiments on the 7-link biped and experiments using simulators with different levels of fidelity. We demonstrate that our proposed approach is able to generalize to different controllers and robot structures and is also robust to simulation inaccuracies.

\subsection{Hardware Experiments on the ATRIAS Robot}
In this section we describe experiments conducted on the ATRIAS robot, described in Section \ref{subsec:atrias}. These experiments were conducted around a boom. The cost function used in our experiments is a slight modification of the cost used in~\citep{song2015neural}:
\begin{equation}
    \label{eq:hdw_cost}
    cost = 
    \begin{cases}
		100 - x_{fall} , \text{\small{if fall}} \\
		||v_{avg} - v_{tgt}||, \text{\small{if walk}}\\
	\end{cases}
\end{equation}
where $x_{fall}$ is distance covered before falling, $v_{avg}$ is average speed per step and $v_{tgt}$ contains target velocity profile, which can be variable. This cost function heavily penalizes falls, and encourages walking controllers to track target velocity. 

\begin{figure*}[t]
\centering
\includegraphics[width=0.725\textwidth]{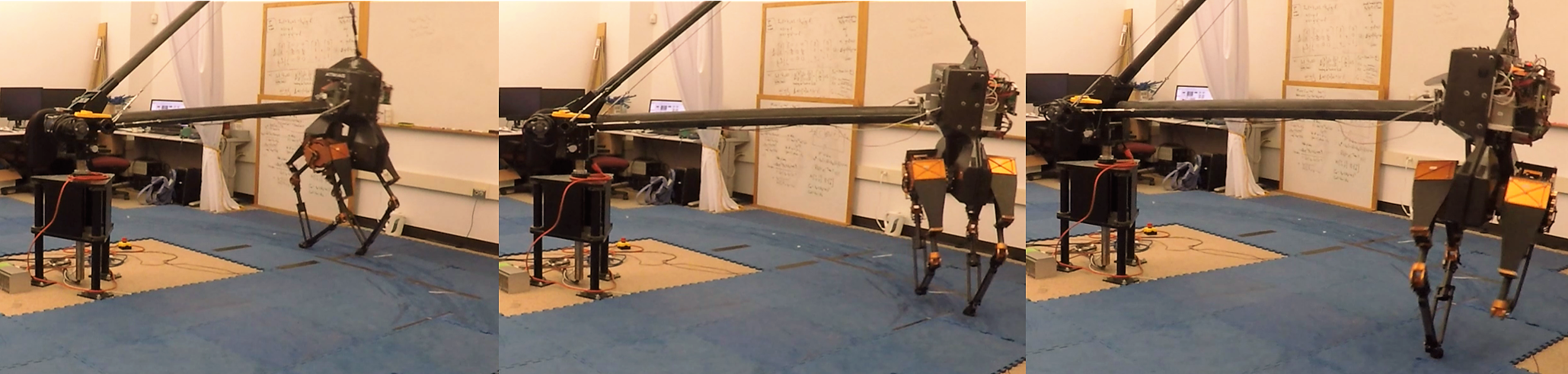}
\caption{\small{ATRIAS during BO with DoG-based kernel (video: \url{https://youtu.be/hpXNFREgaRA})}}
\label{fig:bo_runs_atrias_hw_slides}
\vspace{-5mm}
\end{figure*}

We do multiple runs of each algorithm on the robot. Each run typically consists of 10 experiments on the robot. Hence 3 runs for one algorithm involve 30 robot trials. Each robot trial is designed to be between $30s$ to a minute long and the robot needs to be reset to its ``home" position between trials. While this might not appear to be very time consuming, often parts of the robot malfunction between trials and repairs need to be done, especially when sampling unstable controllers. We try our best to keep the robot performance consistent across the different algorithms being compared.  

We will present two sets of hardware experiments in the following sections. First we present experiments with the DoG-based kernel on the 5 and 9 dimensional controllers introduced in Section \ref{subsec:raibert_cont}. In these experiments from our work in~\citet{rai2017bayesian}, the inertial measurement unit (IMU) of the robot had been damaged, and we replaced it with external boom sensors. While these sensors give all the required information, they are lower resolution than the IMU, leading to noisier readings and larger time delays. This makes these experiments especially challenging. In our second set of experiments, we optimize a 9-dimensional controller using a Neural Network based kernel on hardware. In this new set of experiments the IMU had been fixed, leading to better state estimation on the robot. As a result, the behavior of the robot was slightly changed, and we re-conducted experiments for the baseline for this setting. The baseline performed slightly better than the first set of experiments, as can be expected as a result of improved sensing on the robot.

\subsubsection{Experiments with a 5-dimensional controller and DoG-based kernel}

In our first set of experiments on the robot, we investigated optimizing the 5-dimensional controller from \mbox{Section~\ref{subsec:raibert_cont}}. For these experiments we picked a challenging variable target speed profile:
$0.4 m/s \text{ (15 steps) - } 1.0 m/s \text{ (15 steps) - } 0.2 m/s \text{ (15 steps)}$ $ \text{- } 0 m/s \text{ (5 steps)}$. The controller was stopped after the robot took 50 steps.

\begin{figure}[t]
\begin{subfigure}[t]{0.47\textwidth}
\centering
\includegraphics[width=1.0\textwidth]{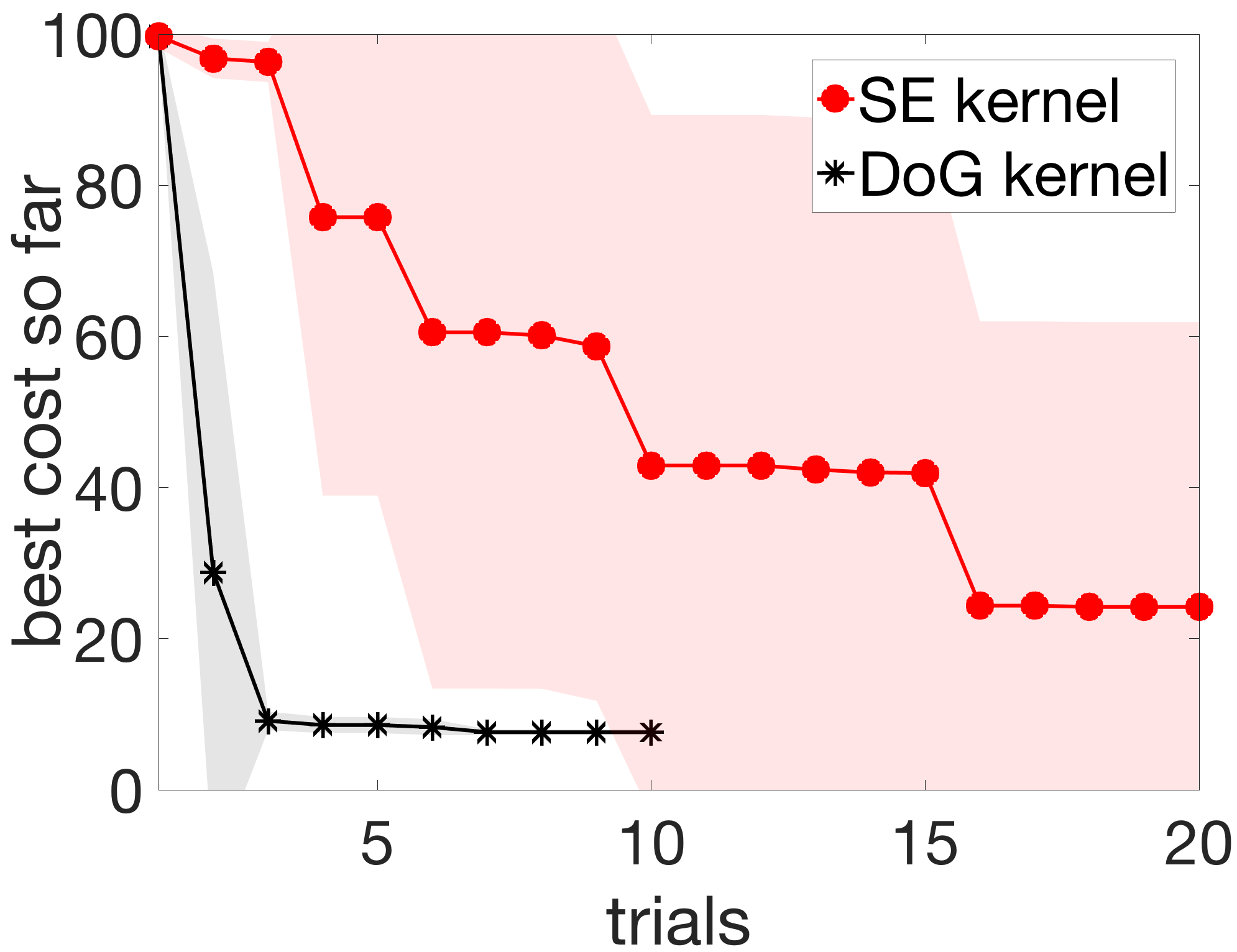}
\caption{\small{BO for 5D controller. BO with SE finds walking points in 4/5 runs within 20 trials. BO with DoG-based kernel finds walking points in 5/5 runs within 3 trials.}}
        \vspace{-5mm}
\label{fig:hw_raibert_5d}
\end{subfigure}
\hspace{10px}
\begin{subfigure}[t]{0.47\textwidth}
\centering
\includegraphics[width=1.0\textwidth]{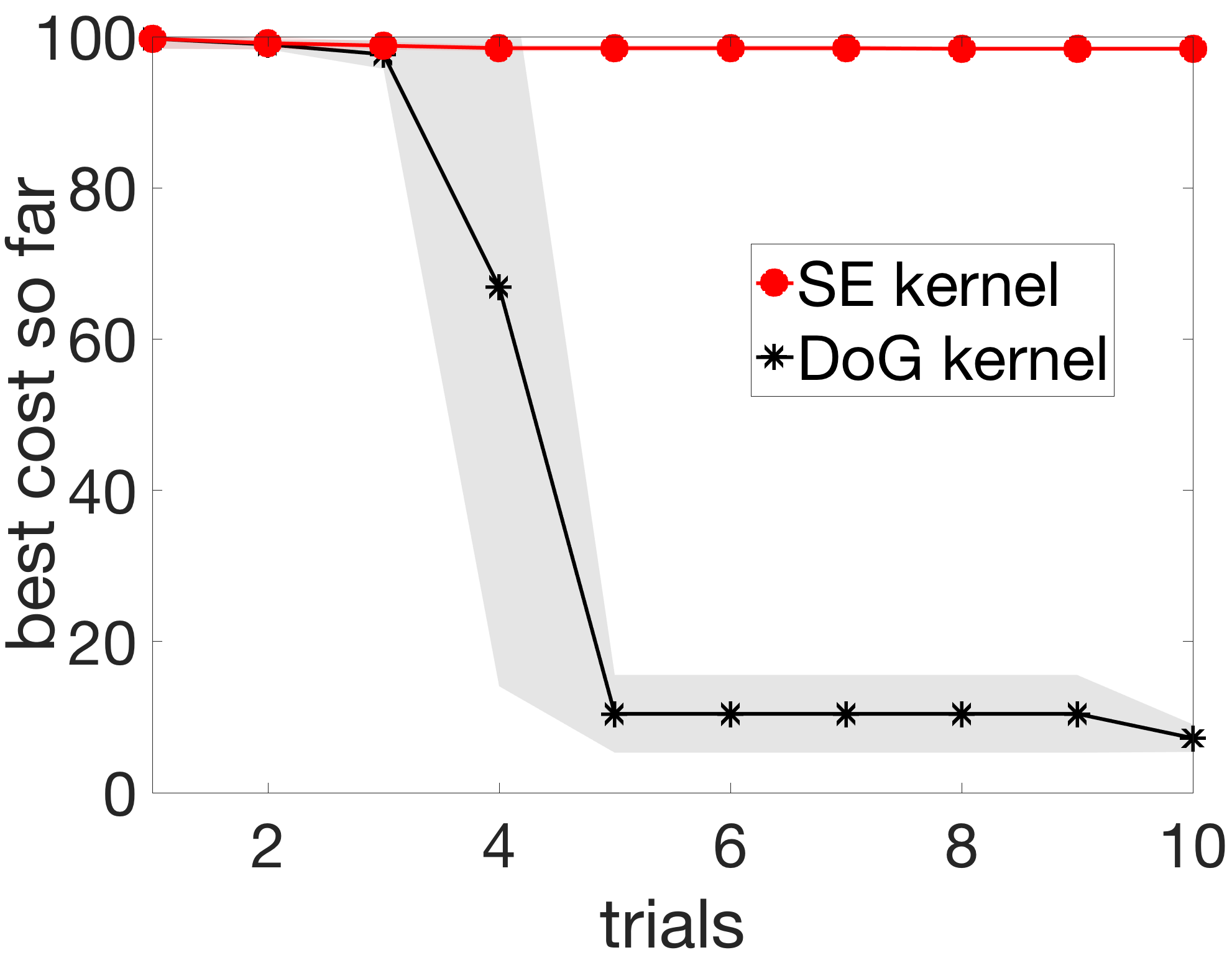}
\caption{\small{BO for 9D controller. BO with SE doesn't find walking points in 3 runs. BO with DoG-based kernel finds walking points in 3/3 runs within 5 trials.}}
\vspace{-2mm}
\label{fig:hw_raibert_9d}
\end{subfigure}
\caption{\small{BO for 5D and 9D controller on ATRIAS robot hardware. Plots show mean best cost so far. Shaded region shows $\pm$ one standard deviation. Re-created from~\citet{rai2017bayesian}.}}
\end{figure}

To evaluate the difficulty of this setting, we sampled 100 random points on hardware. 10\% of these were found to walk. In contrast, in simulation the success rate of random sampling was $\approx$27.5\%. This indicates that the simulation was easier, which could be potentially detrimental to algorithms that rely heavily on simulation, because a large portion of controllers that walk in simulation fall on hardware. Nevertheless, using a DoG-based kernel offered significant improvements over a standard SE kernel, as shown in  Figure~\ref{fig:hw_raibert_5d}. 

We conducted 5 runs of each – BO with DoG-based kernel and BO with SE, 10 trials for DoG-based kernel per run, and 20 for SE kernel. In total, this led to 150 experiments on the robot (excluding the 100 random samples). BO with DoG-based kernel finds walking points in 100\% of runs within 3 trials. In comparison, BO with SE found walking points in 10 trials in 60\% runs, and in 80\% runs in 20 trials (Figure \ref{fig:hw_raibert_5d}).

\subsubsection{Experiments with a 9-dimensional controller and DoG-based kernel}
\label{subsec:dog-9d}
Our next set of experiments optimized the 9-dimensional controller from Section \ref{subsec:raibert_cont}. First we sampled 100 random points for the variable speed profile described above, but this led to no walking points. To ensure that we have a reasonable baseline we decided to simplify the speed profile for this setting: $0.4 m/s$ for $30$ steps. We evaluated 100 random points on hardware, and 3 walked for the easier speed profile. In comparison, the success rate in simulation is 8\% for the tougher variable-speed profile, implying an even greater mismatch between hardware and simulation than the 5-dimensional controller. 
Part of the mismatch can be attributed to the lack of IMU in these experiments. In the 9-dimensional controller, the desired CoM height as well as the feedback gains for this height are optimized. Without the IMU, our system does not have a good estimation of vertical height of the CoM, except through kinematics, leading to poor control authority. 
However, the IMU on ATRIAS is a very expensive fiber-optic IMU that is not commonly used on humanoid robots, and most robots use simple state estimation methods. So, this is a common setting for humanoid robots, even if it presents a challenge for the optimization methods.

We conducted 3 runs of BO with DoG-based kernel and BO with SE, 10 trials for DoG-based kernel per run, and 10 for SE. In total, this led to 60 experiments on the hardware (excluding the random sampling). BO with DoG-based kernel found walking points in 5 trials in 3/3 runs. BO with SE did not find any walking points in 10 trials in all 3 runs. These results are shown in Figure~\ref{fig:hw_raibert_9d}.

Based on these results, we concluded that BO with DoG-based kernel was indeed able to extract useful information from simulation and speed up learning on hardware, even when there was mismatch between simulation and hardware.

\subsubsection{Experiments with a 9-dimensional controller and NN-based kernel}

\begin{wrapfigure}{r}{0.4\textwidth}
\vspace{-15px}
\centering
\includegraphics[width=0.4\textwidth]{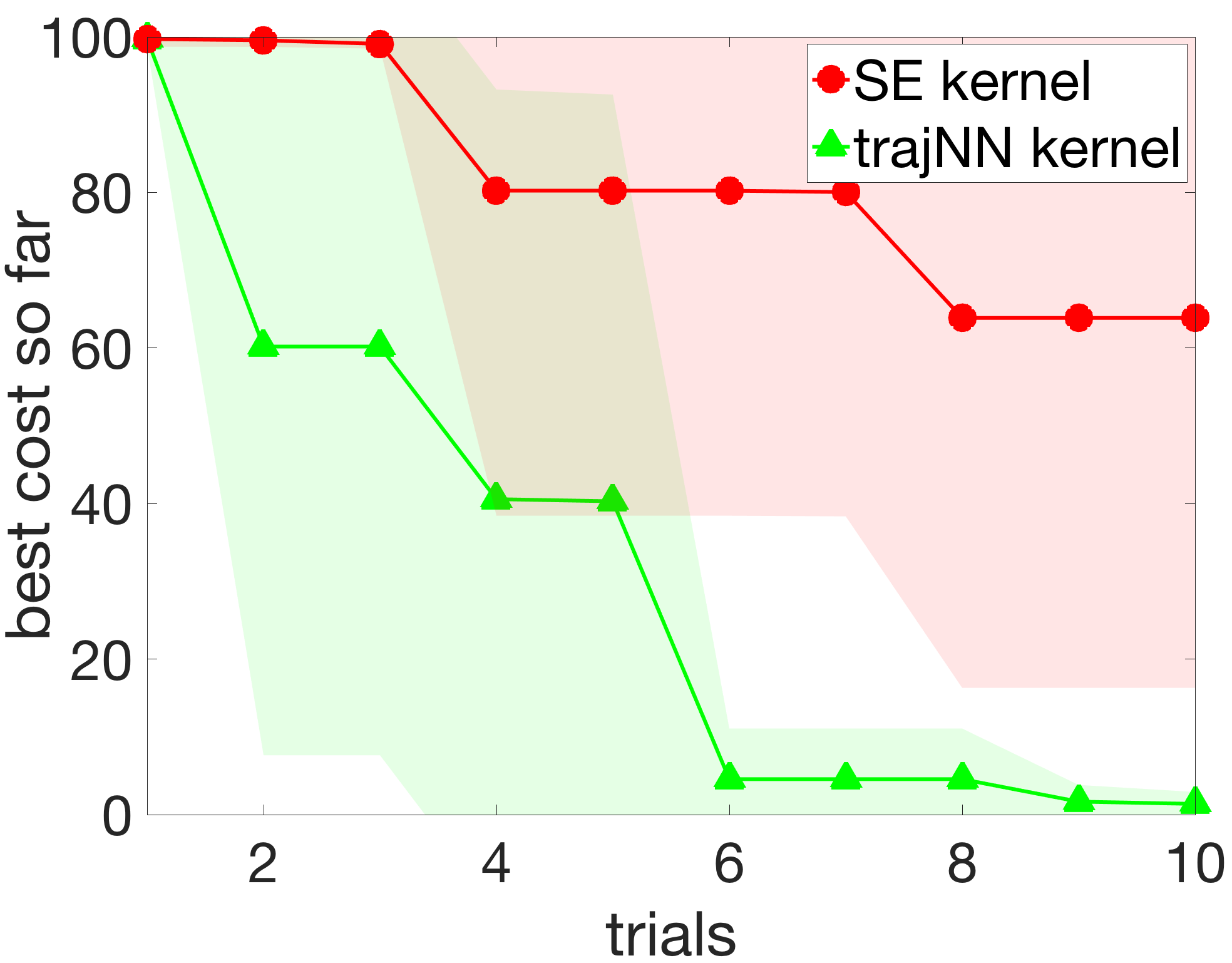}
\caption{\small{BO for 9D controller on ATRIAS robot hardware.}}
\label{fig:9d_hw_nn}
\vspace{-5mm}
\end{wrapfigure}

In the next set of experiments, we evaluated performance of the NN-based kernel described in Section~\ref{subsec:proposed_traj_nn}. We optimize the 9-dimensional controller from Section \ref{subsec:raibert_cont}. 

The target of hardware experiments was to walk for 30 steps at $0.4m/s$, similar to Section \ref{subsec:dog-9d}. However, by these experiments the IMU had been re-installed on the robot. 

We observed that the SE performance improved, even though starting from the same random samples, hyper-parameter setting and speed profile. We attribute this change to a better estimation and control of the CoM vertical height.

Figure~\ref{fig:9d_hw_nn} shows comparison of BO with NN-based kernel and SE kernels. We conducted 5 runs of both algorithms with 10 trials in each run, leading to a total of 100 robot trials. BO with the NN-based kernel found walking points in all 5 runs within 6 trials, while BO with SE kernel only found walking points in 2 of 5 runs in 10 trials.
Hence, even without explicit hand-designed domain knowledge, like the DoG-based kernel, the NN-based kernel is able to extract useful information from simulation and successfully guide hardware experiments.

%% file: biped7link_experiments.tex
\subsection{Simulation Experiments on a 7-link Biped}
\label{sec:biped_7link_experiments}
In this section, we discuss simulation experiments with a 16-dimensional Neuromuscular controller (Section~\ref{sec:nmm_cont}) on a 7-link biped model. These experiments, first reported in ~\citet{antonova2017deep}, also demonstrate the cost-agnostic nature of our approach by optimizing two very different costs.

\begin{figure}[t]
\centering
\vspace{-3px}
\begin{subfigure}[t]{0.48\textwidth}
\centering
\vspace{-5px}
\includegraphics[width=0.9\textwidth]{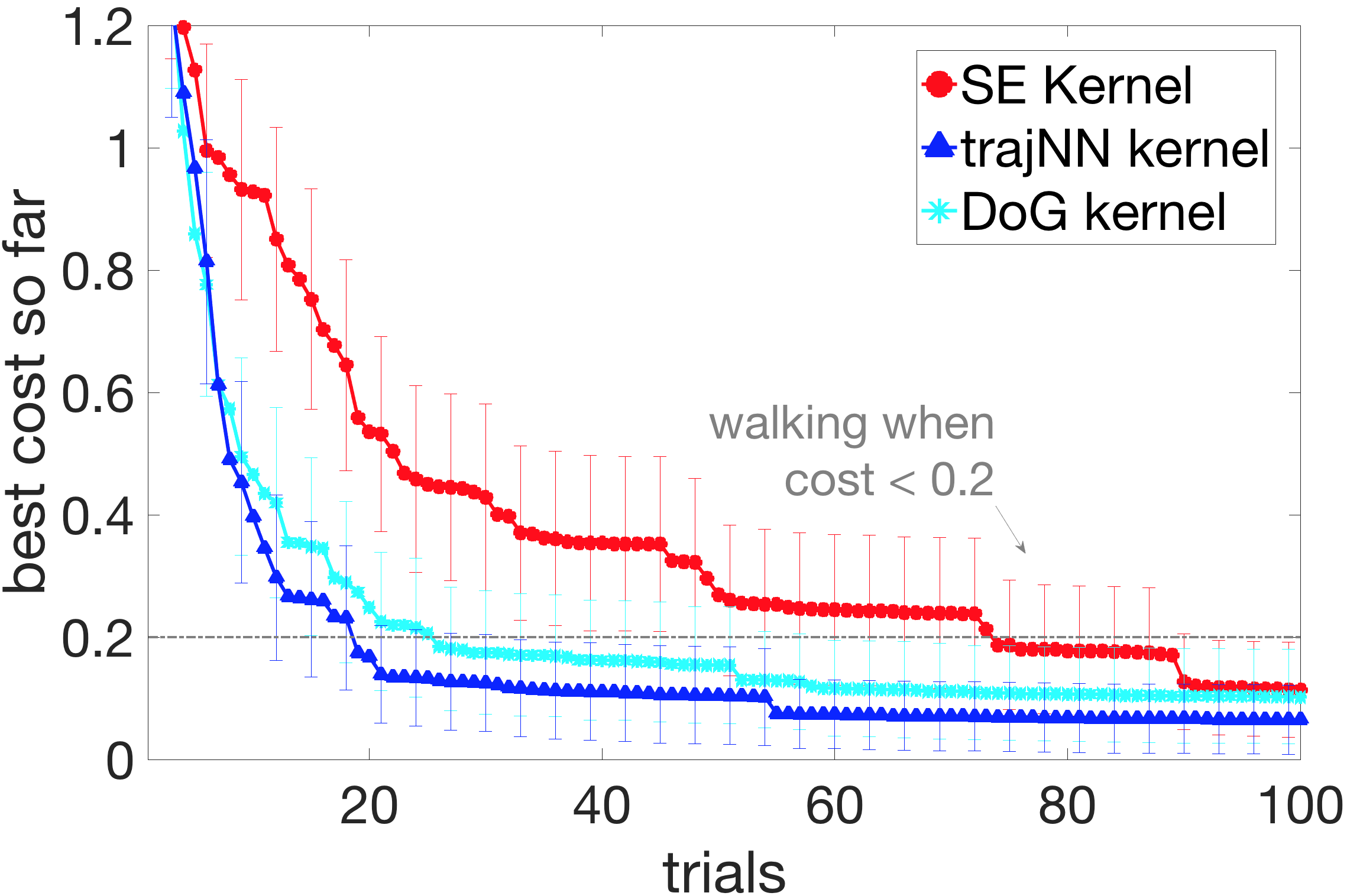}
\caption{\small{Using smooth cost from Equation~\ref{eq:cost_smooth}.}}
\label{fig:smooth_cost_bo_runs}
\end{subfigure}
\quad
\begin{subfigure}[t]{0.48\textwidth}
\centering
\vspace{-5px}
\includegraphics[width=0.9\textwidth]{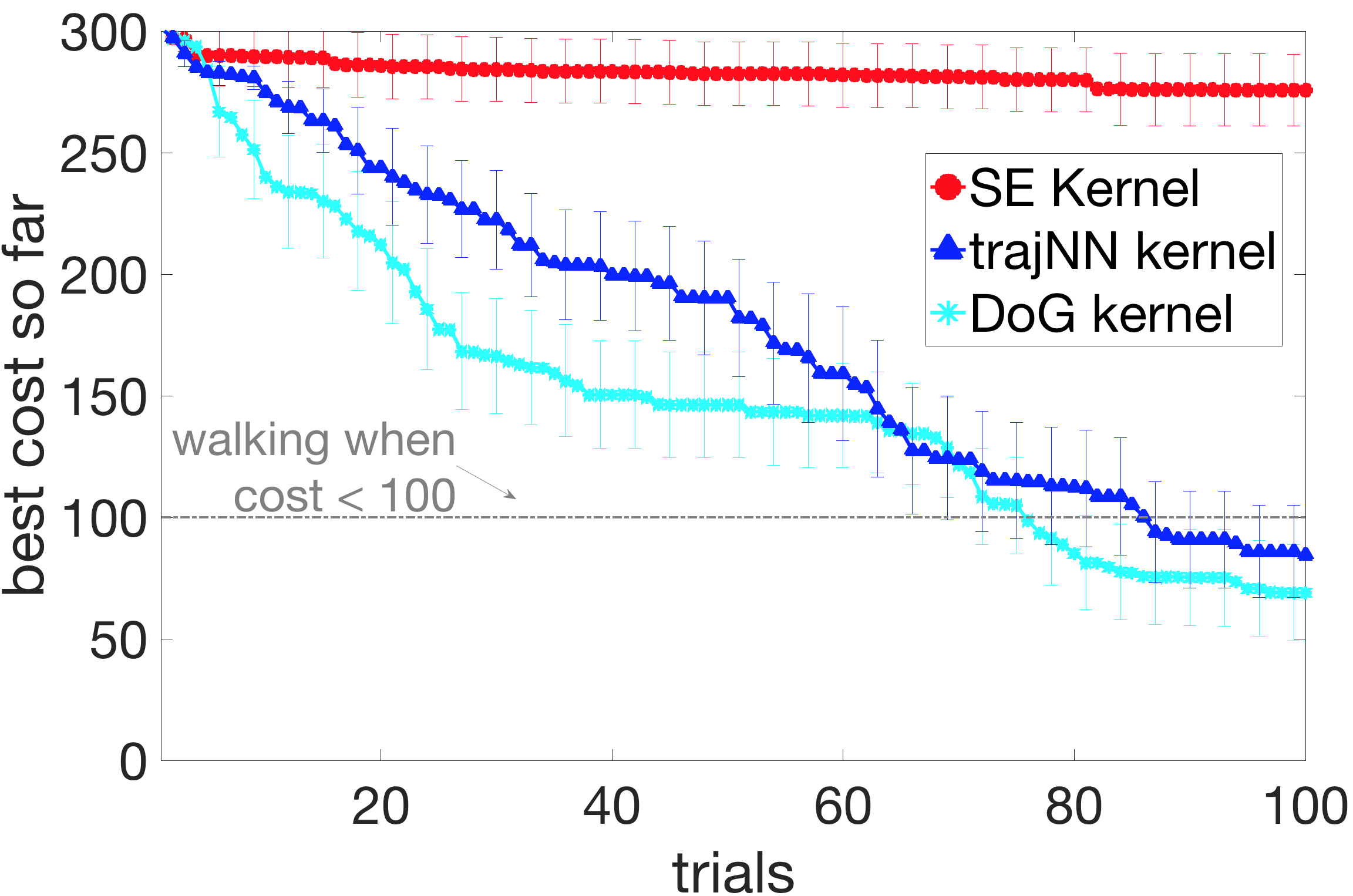}
\caption{\small{Using non-smooth cost from Equation~\ref{eq:cost_nonsmooth}.}}
\label{fig:nonsmooth_cost_bo_runs}
\end{subfigure}
\caption{\small{BO for the Neuromuscular controller. 
\textit{trajNN} and \textit{DoG} kernels were constructed with undisturbed model on flat ground.
BO is run with mass/inertia disturbances on different rough ground profiles.
Plots show means over 50 runs, 95\% CIs. Re-created from~\citet{antonova2017deep}.}}
\label{fig:bo_runs_nm}
\vspace{-10px}
\end{figure}

Figure~\ref{fig:bo_runs_nm} shows BO with DoG-based kernel, NN-based kernel and SE kernel for two different costs from prior literature. The first cost promotes walking further and longer before falling, while penalizing deviations from the target speed~\citep{rai2016sample}:
\begin{equation}
cost_{smooth} = 1/(1+t) + 0.3/(1+d) + 0.01(s-s_{tgt}),
\label{eq:cost_smooth}
\end{equation}
where $t$ is seconds walked, $d$ is the final CoM position, $s$ is speed and $s_{tgt}$ is the desired walking speed ($1.3m/s$ in our case). The second cost function is similar to the cost used in Section \ref{sec:hw_experiments}. It penalizes falls explicitly, and encourages walking at desired speed and with lower cost of transport:
\vspace{-3px}
\begin{equation}
cost_{non\text{-}smooth} = 		
    \begin{cases}
		300 - x_{fall} , \text{\small{if fall}} \\
		100 ||v_{avg} - v_{tgt}|| + c_{tr}, \text{\small{if walk}}\\
	\end{cases}
\label{eq:cost_nonsmooth}
\end{equation}
where $x_{fall}$ is the distance covered before falling, $v_{avg}$ is the average speed of walking, $v_{tgt}$ is the target velocity, and $c_{tr}$ captures the cost of transport. The changed constant is to account for a longer simulation time.

Figure~\ref{fig:smooth_cost_bo_runs} shows that the NN-based kernel and the DoG-based kernel offer a significant improvement over BO with the SE kernel in sample efficiency when using the $cost_{smooth}$, with more than 90\% of runs achieving walking after 25 trials. BO with the SE kernel takes 90 trials to get 90\% success rate. 
Figure~\ref{fig:nonsmooth_cost_bo_runs} shows that similar performance by the two proposed approaches is observed on the non-smooth cost. 
With the NN-based kernel, 70\% of the runs find walking solutions after 100 trials, similar to the DoG-based kernel. However, optimizing non-smooth cost is very challenging for BO with the SE kernel: a walking solution is found only in 1 out of 50 runs after 100 trials.

We attribute the difference in performance of the SE kernel on the two costs to the nature of the costs. If a point walks some distance $d$, Equation \ref{eq:cost_smooth} reduces in terms of $\tfrac{1}{d}$ and Equation \ref{eq:cost_nonsmooth} reduces by $-d$. A sharper fall in the first cost causes BO to exploit around points that walk at some distance, finding points that walk forever. BO with the second cost continues to explore, as the signal is too weak. However the success of both NN-based and DoG-based kernels on both costs shows that the same kernel can indeed be used for optimizing multiple costs robustly, without any further tuning needed. This is important because often the cost has to be changed based on the outcome of the optimization, and it would be impractical to recreate the kernel for each of these costs.

%% file: mismatch_experiments.tex
\subsection{Experiments with Increasing Simulation-Hardware Mismatch}
\label{subsec:mismatch_experiments}

\begin{wrapfigure}{r}{0.43\textwidth}
\vspace{-15px}
\centering
\includegraphics[width=0.43\textwidth]{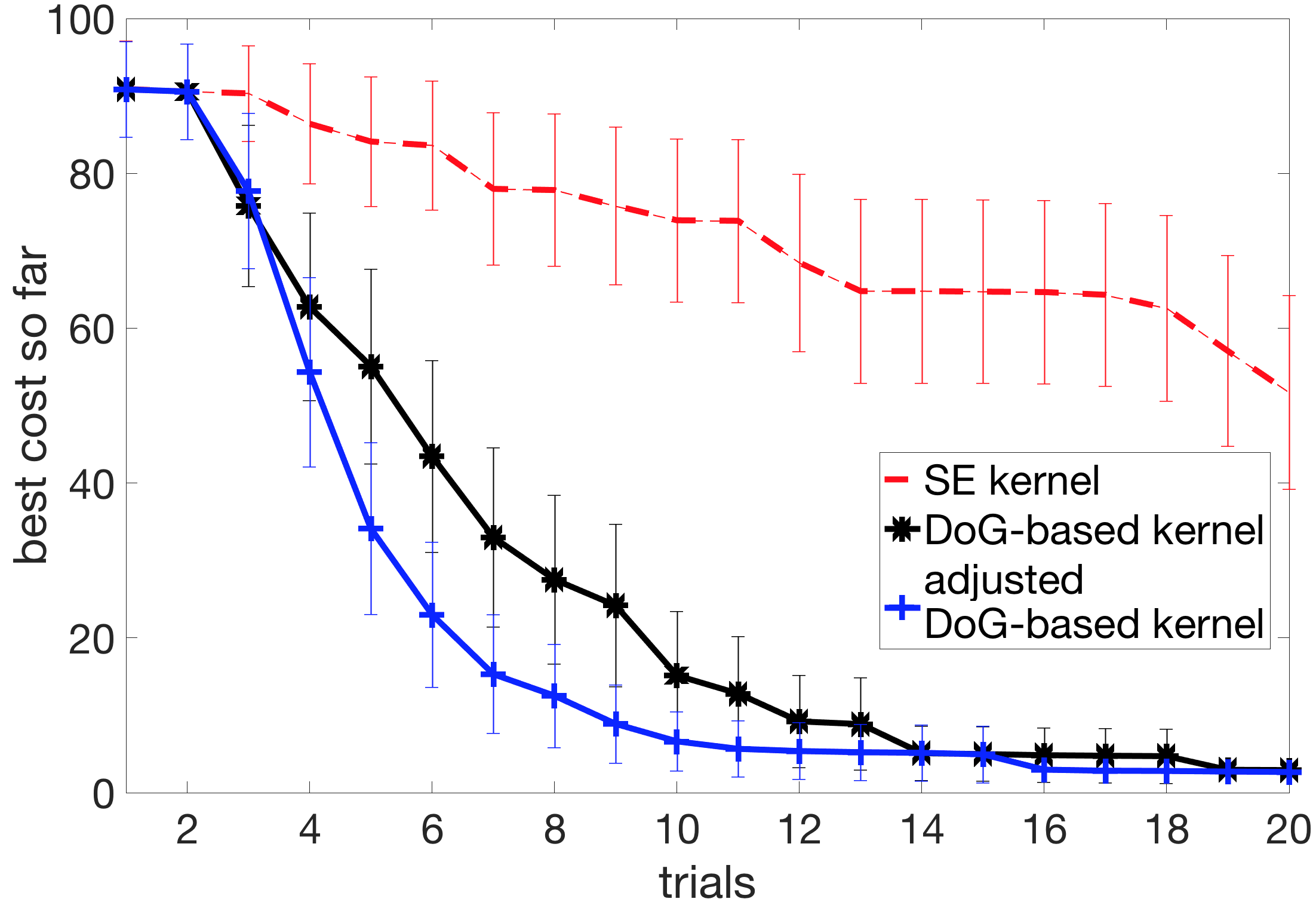}
\caption{\small{BO for 50d controller on original ATRIAS simulation \citep{rai2017bayesian}.}}
\label{fig:50d_sim}
\vspace{-2mm}
\end{wrapfigure}

In this section, we describe our experiments with increasing simulation-hardware mismatch and its effect on approaches that use information from simulation during hardware optimization. The quality of information transfer between simulation and hardware depends not only on the mismatch between the two, but also on the controller used. For a robust controller, small dynamics errors would not cause a significant deterioration in performance, while for a sensitive controller this might be much more detrimental. There is still an advantage to studying such a sensitive controller, as it might be much more energy efficient and versatile. In our experiments, the 50-dimensional VNMC described in Section \ref{subsec:VNMC_cont} is capable of generating very efficient gaits but is sensitive to modelling errors. Figure \ref{fig:50d_sim} shows the performance of the DoG-based and adjusted DoG-based kernel on the original high-fidelity simulator. While both methods find walking points in 20 trials, adjusted-DoG performs better. There is mismatch even between short $5s$ and long $30s$ simulations for this controller. This mismatch is compensated by the adjusted-DoG kernel.

In the rest of this section, we provide experimental analysis of settings with increasing simulated mismatch and their effect on optimization of the 50-dimensional VNMC. We compare several approaches that improve sample-efficiency of BO and investigate if the improvement they offer is robust to mismatch between the simulated setting used for constructing kernel/prior and the setting on which BO is run.

\begin{figure}[t]
\begin{subfigure}[t]{0.47\textwidth}
\centering
\includegraphics[width=1.0\textwidth]{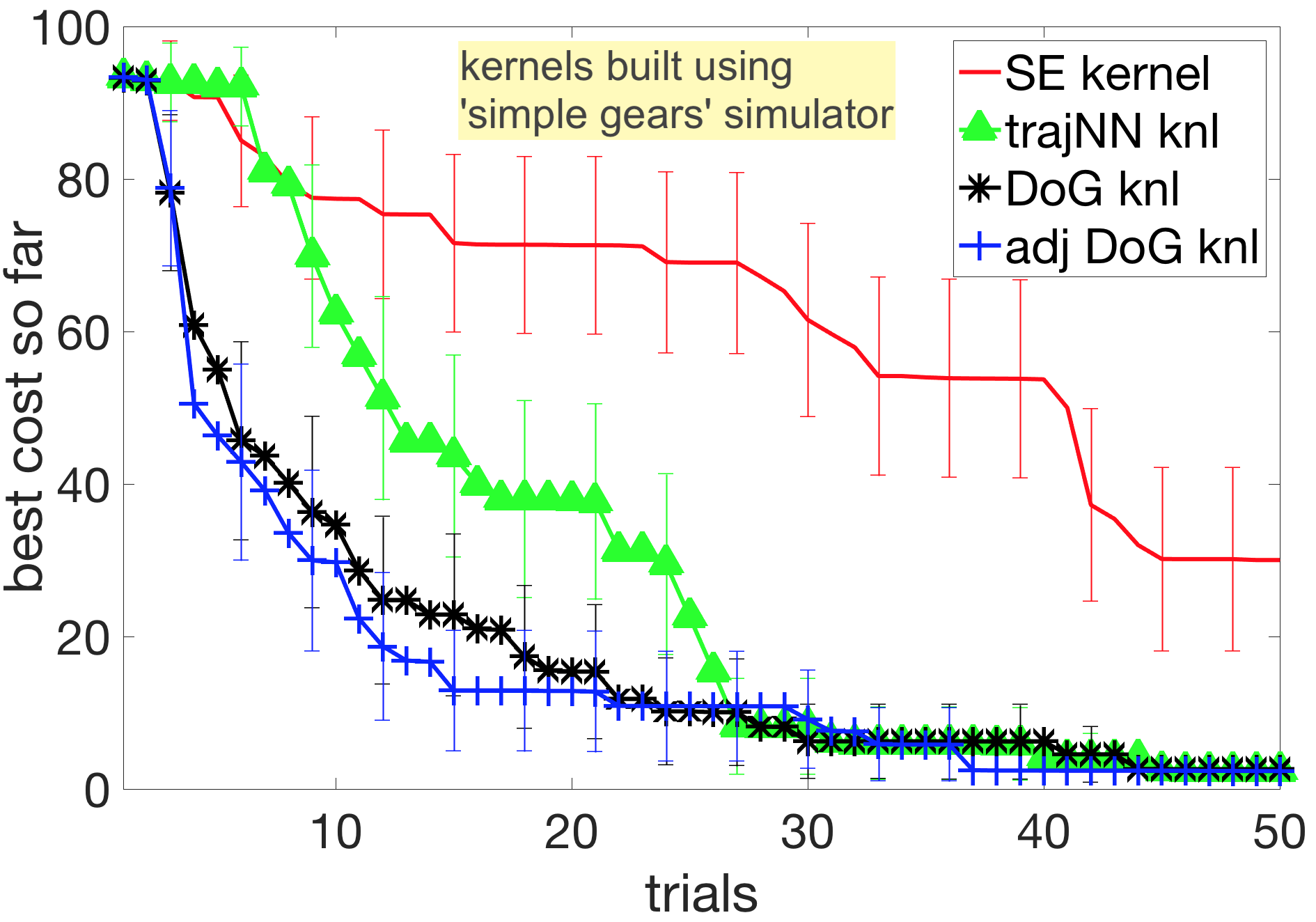}
\vspace{-2mm}
\caption{\small{Informed kernels generated using simulator with simplified gear dynamics.}}
        \vspace{-5mm}
\label{fig:compare_gear_dynamics}
\end{subfigure}
\hspace{10px}
\begin{subfigure}[t]{0.47\textwidth}
\centering
\includegraphics[width=1.0\textwidth]{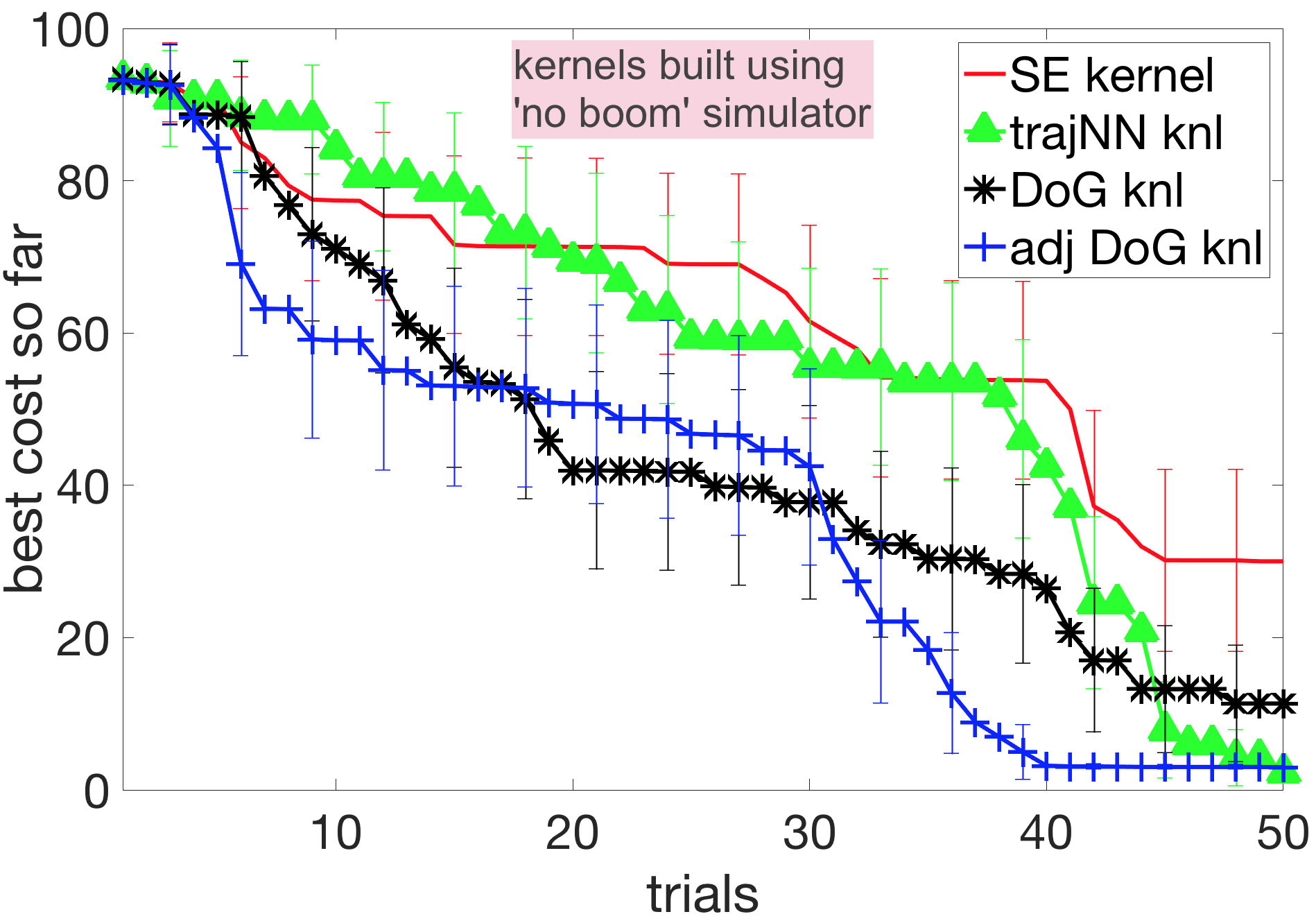}
\vspace{-2mm}
\caption{\small{Informed kernels generated using simplified gear dynamics, without boom model.}}
\label{fig:compare_without_boom}
\end{subfigure}
\vspace{-2mm}
\caption{\small{BO is run on the original simulator. Informed kernels perform well despite significant mismatch, when kernels are generated using simulator with simplified gear dynamics (left). In the case of severe mismatch, when the boom model is also removed, informed kernels still improve over baseline SE (right). Plots show best cost for mean over 50 runs for each algorithm, 95\% CIs.}}
\label{fig:compare_dog}
\end{figure} 
First, we examine the performance of our proposed approaches with informed kernels: $k_{DoG}$, $k_{\text{trajNN}}$ and $k_{DoG_{adj}}$. Figure~\ref{fig:compare_gear_dynamics} shows the case when informed kernels are generated using the simulator with simplified gear dynamics while BO is run on the original simulator. After 50 trials, all runs with informed kernels find walking solutions, while for SE only $70\%$ have walking solutions.

Next, Figure~\ref{fig:compare_without_boom} shows performance of $k_{DoG}$, $k_{\text{trajNN}}$ and $k_{DoG_{adj}}$ when the kernels are constructed using a simulator with simplified dynamics and without a the boom. In this case the mismatch with the original simulator is larger than before and we see the advantage of using adjustment for DoG-based kernel: $k_{DoG_{adj}}$ finds walking points in all runs after 35 trials. $k_{\text{trajNN}}$ also achieves this, but after 50 trials. $k_{DoG}$ finds walking points in $\approx\!90\%$ of the runs after 50 trials. The performance of SE stays the same, as it uses no prior information from any simulator.

This illustrates that while the original DoG-based kernel can recover from slight simulation-hardware mismatch, the adjusted DoG-kernel is required if one expects higher mismatch.  $k_{\text{trajNN}}$ seems to recover from the mismatch, but might benefit from an adjusted version. We leave this to future work.

%------------------------------------------------------------------------------------
\subsubsection{Comparisons of Prior-based and Kernel-based Approaches}
\label{subsec:sim_experiments_prior_cully}
We will classify approaches that use simulation information in hardware optimization as prior-based or kernel-based. Prior-based approaches use costs from the simulation in the prior of the GP used in the BO. This can help BO a lot if the costs between simulation and hardware transfer, and the cost function is fixed. However, in the presence of large mismatch, points that perform well in simulation might fail on hardware. A prior-based method can be biased towards sampling promising points from simulation, resulting in an even poorer performance than methods with no prior. Kernel-based approaches consist of methods that incorporate the information from simulation into the kernel of the GP. These can be sample-inefficient as compared to prior-based method, but less likely to be biased towards unpromising regions in the presence of mismatch. They also easily generalize to multiple costs, so that there is no additional computation if the cost is changed. This is important because a lot of these approaches can take several days of computation to generate the informed kernel. For example, \cite{cully2015robots} report taking 2 weeks on a 16-core computer to generate their map. 

It is possible to also combine both prior-based and kernel-based methods, as in \cite{cully2015robots}. We will classify these as `prior-based' methods, since in our experiments prior outweighs the kernel effects for such cases. In our comparison with \cite{cully2015robots}, we will implement a version with and without the prior points. We do not add a cost prior to BO using DoG-based kernel, as this limits us to a particular cost, and high-fidelity simulators. Since both of these can be major obstacles in real robot experiments, we refrain from doing so.

\begin{figure}[t]
\begin{subfigure}[t]{0.47\textwidth}
\centering
\includegraphics[width=1.0\textwidth]{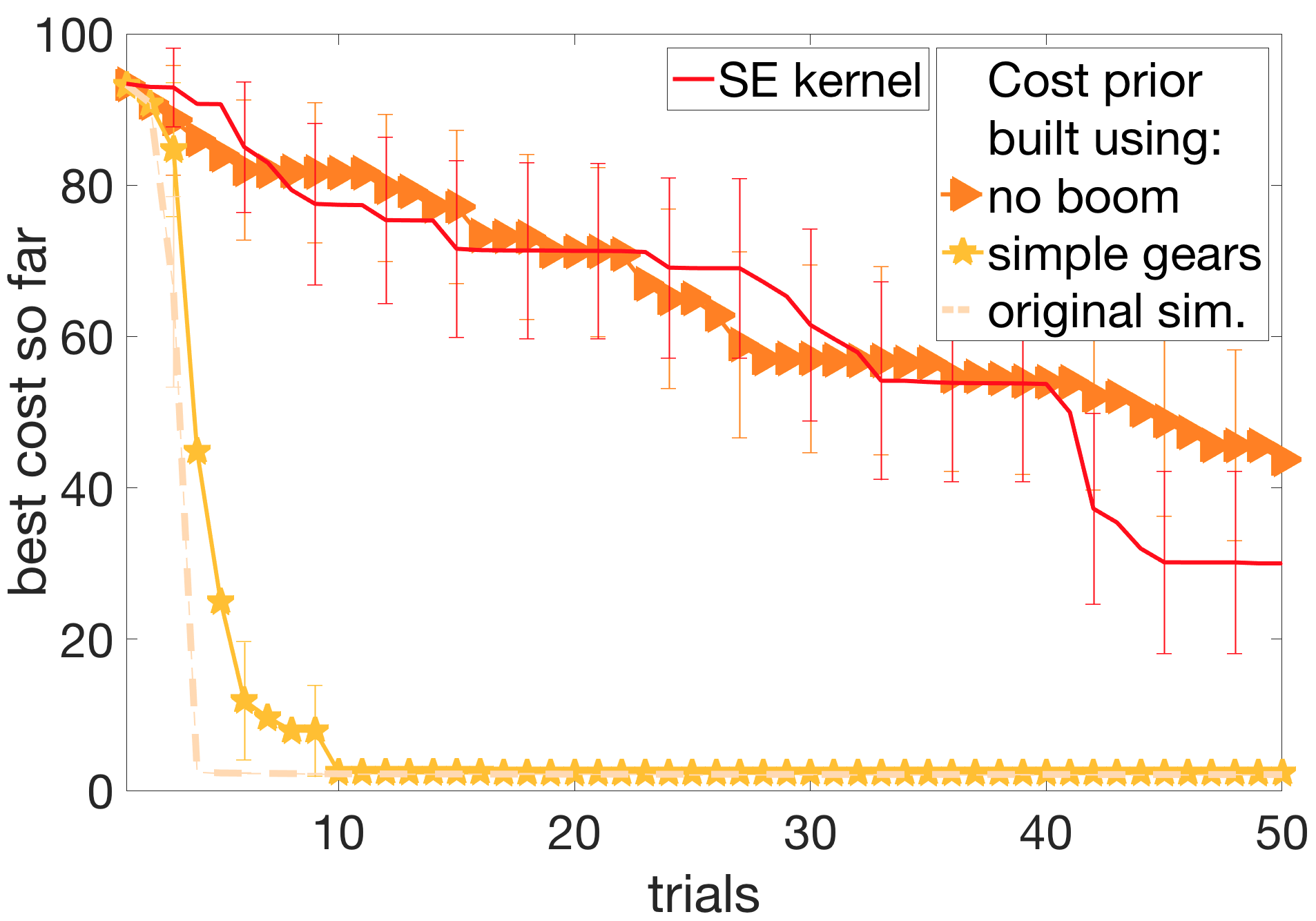}
\caption{\small{BO with cost prior: straightforward approach useful for low-to-medium mismatch; but no improvement if mismatch is severe.}}
\vspace{-5mm}
\label{fig:cost_prior_sim_versions}
\end{subfigure}
\hspace{15px}
\begin{subfigure}[t]{0.47\textwidth}
\centering
\includegraphics[width=1.0\textwidth]{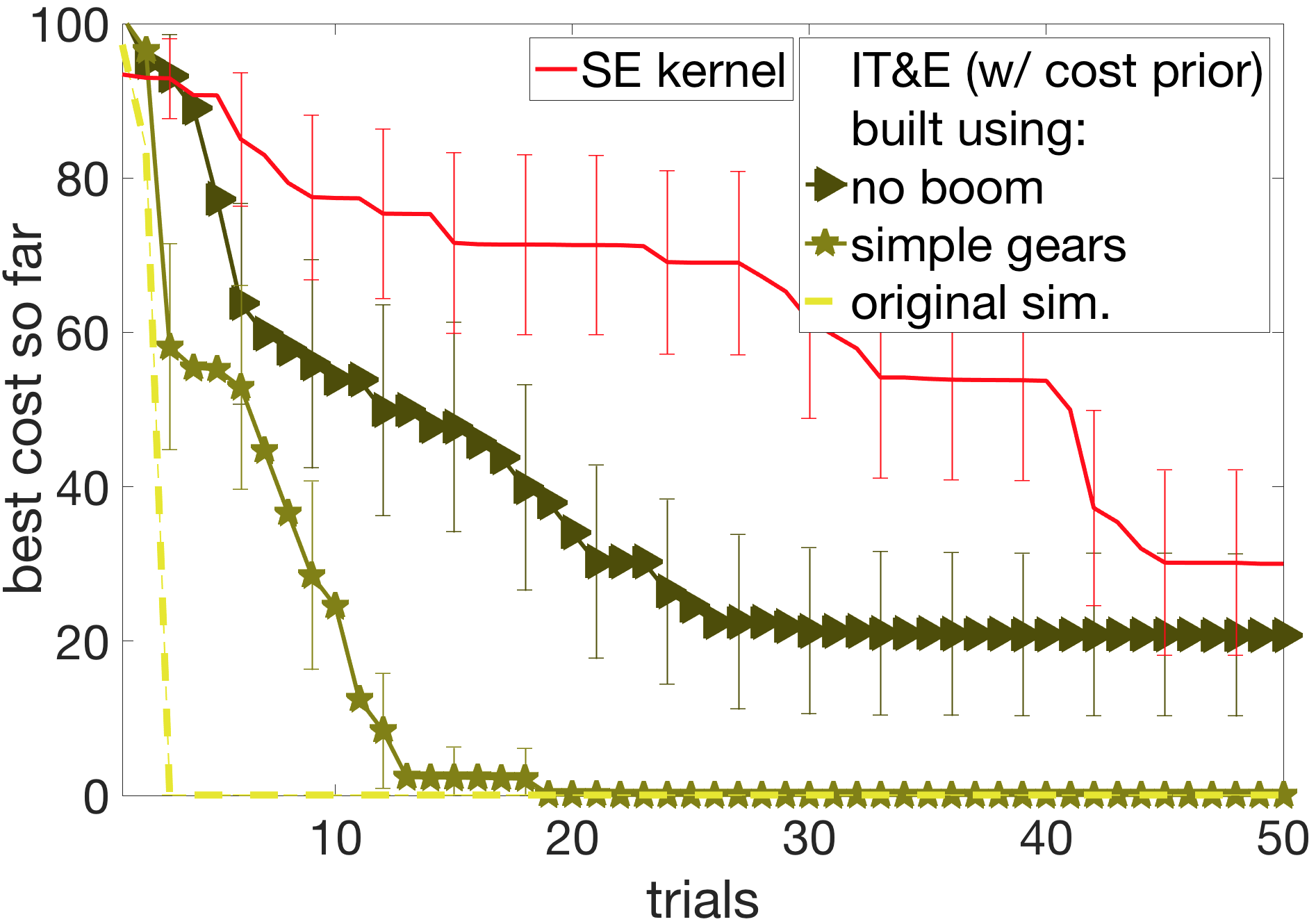}
\caption{\small{Performance of IT\&E algorithm (our implementation of \citet{cully2015robots}, adapted to bipedal locomotion).}}
\vspace{-2mm}
\label{fig:cully_prior_sim_versions}
\end{subfigure}
\caption{\small{BO using prior-based approaches. Mean over 50 runs for each algorithm, 95\% CIs.}}
\label{fig:prior_based_bo}
\end{figure}

Figure~\ref{fig:cost_prior_sim_versions} shows the performance when using simulation cost in the prior during BO. BO with a cost prior created using the original version of the simulator illustrates what would happen in the best case scenario, as optimization is merely a look-up here. When the simulator with simplified gear dynamics is used for constructing the prior, we observe significant improvements over uninformed BO prior. However, when the prior is constructed from simplified gear dynamics and no boom setting, the approach performs slightly worse than uninformed BO. This shows that while an informed prior can be very helpful when created from a simulator close to hardware, it can hurt performance if simulator is significantly different from hardware. 

Next, we discuss experiments with our implementation of Intelligent Trial and Error (IT\&E) algorithm from~\citet{cully2015robots}. This algorithm combines adding a cost prior from simulated evaluations with adding simulation information into the kernel. IT\&E defines a behavior metric and tabulates best performing points from simulation on their corresponding behavior score. The behavior metric used in our experiments is duty-factor of each leg, which can go from 0 to 1.0. We discretize the duty factor into 21 cells of 0.05 increments, leading to a $21\times21$ grid. We collect the 5 highest performing controllers for each square in the behavior grid, creating a $21\times21\times5$ grid. Next, we generate 50 random combinations of a $21\times21$ grid, selecting 1 out of the 5 best controllers per grid cell. Care was taken to ensure that all 5 controllers had comparable costs in the simulator used for creating the map. 
Cost of each selected controller is added to the prior and BO was performed in the behavior space, like in \cite{cully2015robots}.

Figure~\ref{fig:cully_prior_sim_versions} shows BO with IT\&E constructed using different versions of the simulator. \mbox{IT\&E} constructed using simplified gear dynamics simulator is slightly less sample-efficient than the straightforward `cost prior' approach. When constructed with the simulator with no boom, \mbox{IT\&E} is able to improve over uninformed BO. However, it only finds walking points in 77\% of the runs in 50 trials in  this case, as some of the generated maps contained no controllers that could walk on the `hardware'. This is a shortcoming of the IT\&E algorithm, as it eliminates a very large part of the search space and if the pre-selected space does not contain a walking point, no walking controllers can be sampled with BO. This problem could possibly be avoided by using a finer grid, or a different behavior metric. However tuning such hyper-parameters can turn out to be expensive, in computation and hardware experiment time.

\begin{figure}[t]
\begin{subfigure}[t]{0.47\textwidth}
\centering
\includegraphics[width=1.0\textwidth]{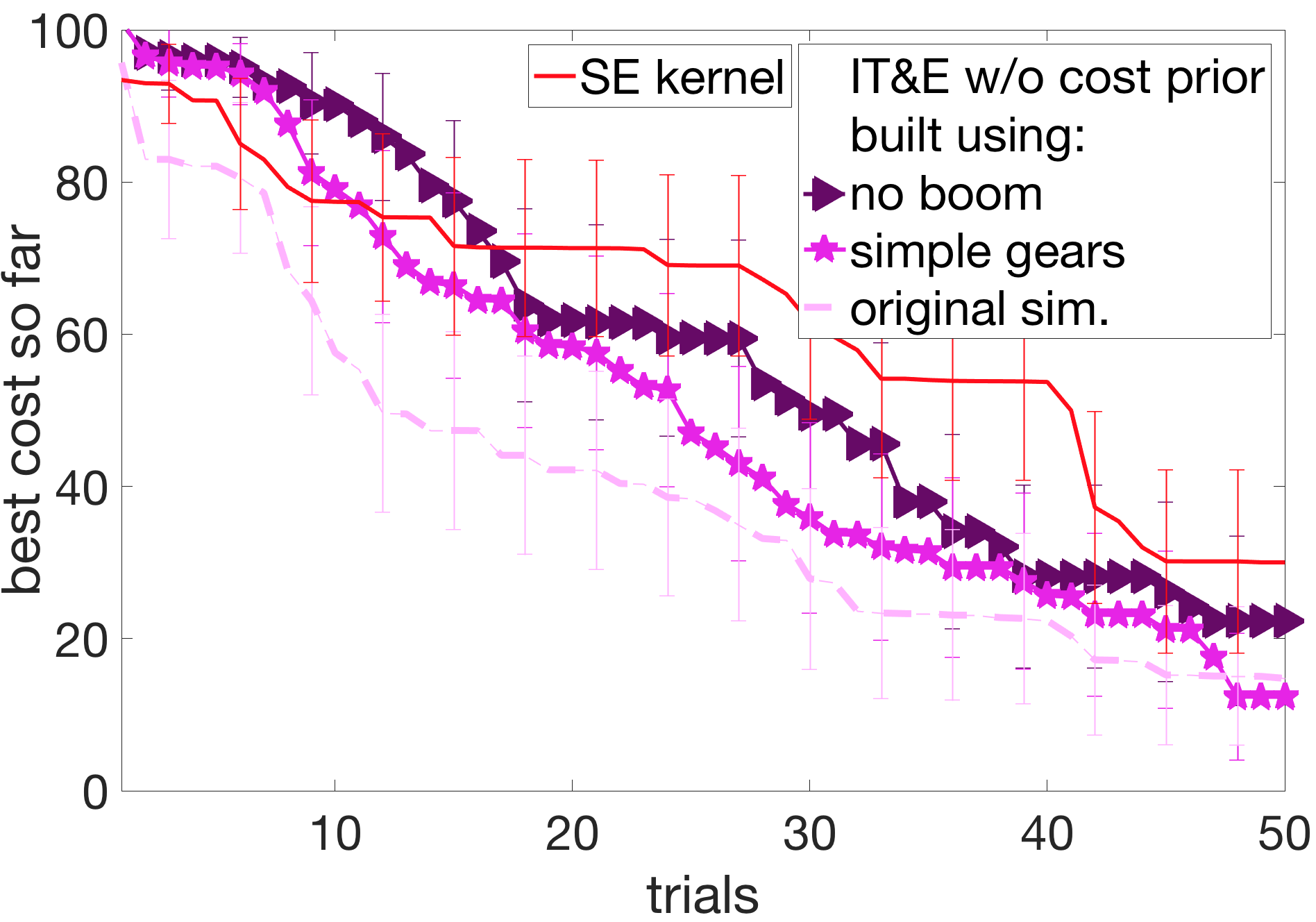}
\caption{\small{BO using our implementation of IT\&E without cost prior (from \citet{cully2015robots}).}}
\vspace{-5mm}
\label{fig:cully_kernel_sim_versions}
\end{subfigure}
\hspace{10px}
\begin{subfigure}[t]{0.47\textwidth}
\centering
\includegraphics[width=1.0\textwidth]{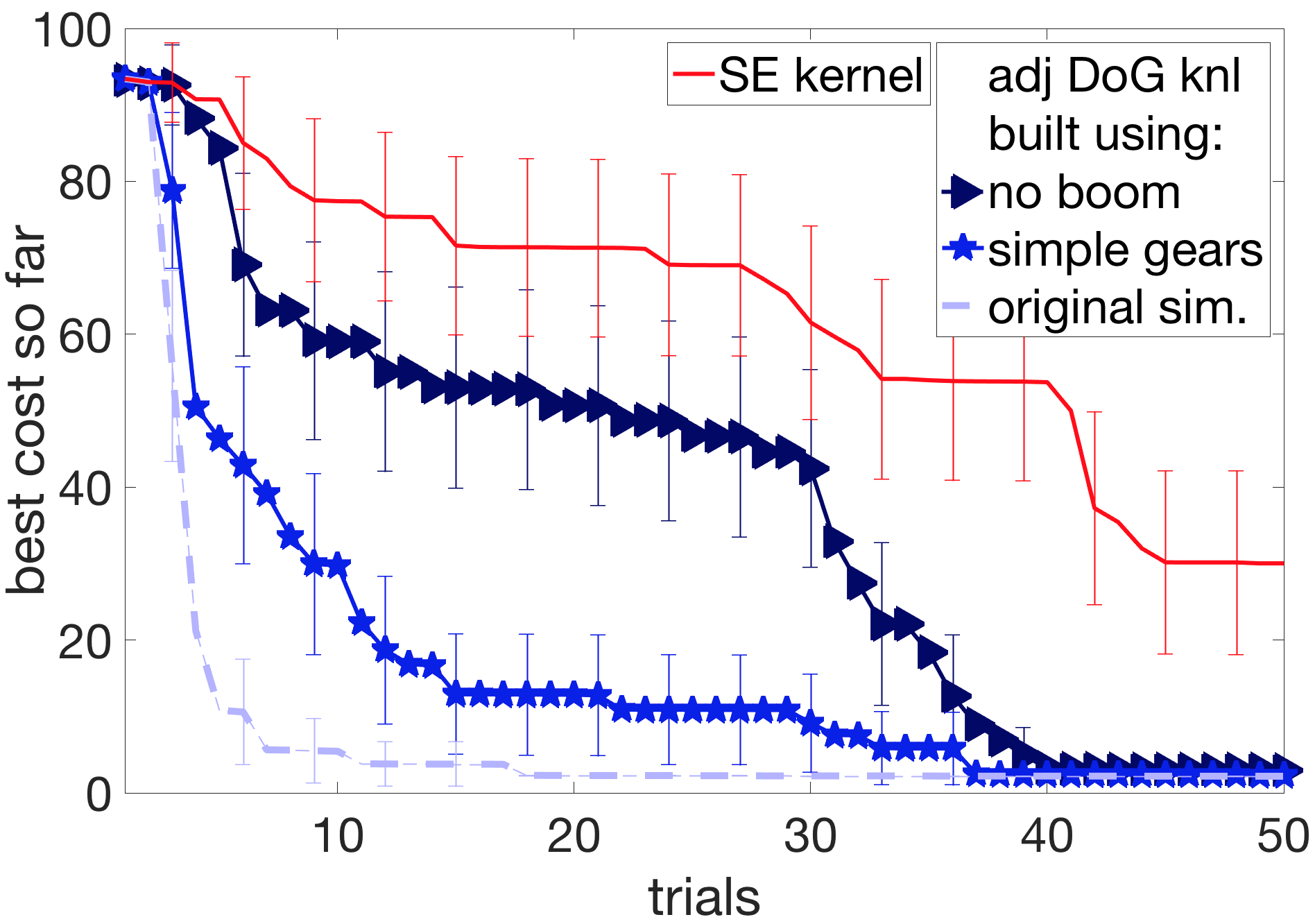}
\caption{\small{BO using $k_{DoG_{adj}}$ constructed from simulators with various levels of mismatch.}}
\vspace{-2mm}
\label{fig:kernel_hwadjust2_sim_versions}
\end{subfigure}
\caption{\small{BO using kernel-based approaches. Mean over 50 runs for each algorithm, 95\% CIs.}}
\label{fig:prior_based_bo}
\end{figure}
To separate the effects of using simulation information in prior mean vs kernel, we evaluated a kernel-only version of \mbox{IT\&E} algorithm. Figure~\ref{fig:cully_kernel_sim_versions} shows these results. It shows that the cost prior is crucial for the success of IT\&E and performance deteriorates without it. Hence, it is not practical to use IT\&E on a cost different than what it was generated for.

Nonetheless, Figure~\ref{fig:compare_dog} showed that BO with adjusted DoG kernel is able to handle both moderate and severe mismatch with kernel-only information, collected in  Figure~\ref{fig:kernel_hwadjust2_sim_versions}.

In summary, we created two simulators with increasing modelling approximations, and studied the effect of using these to aid optimization on the original simulator. We found that while methods that use cost in the prior of BO can be very sample-efficient in low mismatch, their performance worsens as mismatch increases. \mbox{IT\&E} introduced in \cite{cully2015robots} uses simulation information in both prior mean and kernel, and is very sample-efficient in cases of low mismatch. Even with high mismatch, it performs better than just prior-based BO but doesn't find walking controllers reliably. In comparison, adjusted DoG-based kernel performed well in all the tested scenarios. All of this shows that the adjusted DoG-based kernel can reliably improve sample-efficiency of BO even when the mismatch between simulation and hardware is high. We would like to continue working in this direction and explore the usefulness of even simpler simulators in the future.

%% file: conclusion.tex
\section{Conclusion}
\label{sec:conclusion}
In this paper, we presented and analyzed in details our work from \citet{rai2016sample}, \citet{antonova2017deep} and \citet{rai2017bayesian}. These works introduce domain-specific feature transforms that can be used to optimize locomotion controllers on hardware efficiently. The feature transforms project the original controller space into a space where BO can discover promising regions quickly.
We described a transform for bipedal locomotion designed with the knowledge of human walking and a neural network based transform that uses more general information from simulated trajectories. Our experiments demonstrate success at optimizing controllers on the ATRIAS robot. Further simulation-based experiments also indicate potential for other bipedal robots. For optimizing sensitive high-dimensional controllers, we proposed an approach to adjust simulation-based kernel using data seen on hardware. To study the performance of this, as well as compare our approach to other methods, we created a series of increasingly approximate simulators. Our experiments show that while several methods from prior literature can perform well with low simulation-hardware mismatch (sometimes even better than our proposed approach), they suffer when this mismatch increases. In such cases, our proposed kernels with hardware adjustment can yield reliable performance across different costs, simulators and robots.

%% file: appendix_A.tex
\section*{Appendix A: Implementation Details}
\label{apx:appendix_implementation_details}

In this Appendix we provide a summary of data collection and implementation details. Our implementation of BO was based on the framework in~\citet{gardner2014bayesian}. We used Expected Improvement (EI) acquisition function~\citep{mockus1978toward}. We also experimented with Upper Confidence Bound (UCB)~\citep{srinivas2010gaussian}, but found that performance was not sensitive to the choice of acquisition function. Hyper-parameters for BO were initialized to default values: 0 for mean offset, 1.0 for kernel length scales and signal variance, 0.1 for $\sigma_{n}$ (noise parameter). Hyperparameters were optimized using the marginal likelihood (\cite{BOtutorial2016}, Section~V-A). For all algorithms, we optimized hyperparameters after a low-cost controller was found (to save compute resources and avoid premature hyperparameter optimization).

\begin{table}[h!]
\centering
\ra{1.3}
\scriptsize{
\begin{tabular}{ lccccc } 
\toprule
Kernel type & Controller dim & \# Sim points & Sim duration & Kernel dim & Features in kernel\\
\midrule
$k_{DoG}$ & 5 & 20K & 3.5s & 1 & $score_{DoG}$ \\ 
          & 9 & 100K & 5s & 1 & $score_{DoG}$ \\ 
     & 50 & 200K & 5s & 1 & $score_{DoG}$ \\ 
\hline
$k_{\textit{trajNN}}$ & 9 & 100K & 5s & 4 & $t_{walk}$, $x_{end}$, $\theta_{avg}$, $v_{x,avg}$\\ 
                      & 16 & 100K & 5s & 8 &
                $t_{walk}$, $x_{end}$, $\theta_{end}$, $v_{x,end}$\\
                & & & & & $c_{\tau}$,  $y_{end}$,  $v_{y,end}$, $\dot{\theta}_{end}$ \\ 
                      & 50 & 200K & 5s & 13 & $t_{walk}$, $x_{end}$, $c_{\tau}$, $\pmb{\textit{traj}}_{x}$, $\pmb{\textit{traj}}_{\theta}$ \\ 
\bottomrule
\end{tabular}
}
\caption{Simulation Data Collection Details. $score_{DoG}$ was described in Section \ref{subsec:proposed_dog_transform}. For $k_{\textit{trajNN}}$: $t_{walk}$ is time walked in simulation before falling, $x_{end}$ and $y_{end}$ are the $x$ and $y$ positions of Center of Mass (CoM) at the end of the short simulation, $\theta$ is the torso angle, $\dot{\theta}$ is the torso velocity, $v$ is the CoM speed ($v_{x}$ is the horizontal and $v_y$ is the vertical component), $c_{\tau}$ is the squared sum of torques applied; $\pmb{\textit{traj}}_{x}$, $\pmb{\textit{traj}}_{\theta}$ denote vectors with mean CoM and $\theta$ measurements every second.}
\label{tbl:kernel_details}
\end{table}

Our choice of SE kernel as the baseline for BO was due to its widespread use. The SE kernel belongs to a broader class of Mat\'ern kernels. In some applications, carefully choosing the parameters of Mat\'ern kernel could improve performance of BO. However, Mat\'ern kernels are stationary: $k(\pmb{x}_i, \pmb{x}_j)$ depend only on $r\!=\!\pmb{x}_i\!-\!\pmb{x}_j$ for all $\pmb{x}_{i,j}$. Our approach seeks to build kernels that remove this limitation in a manner informed by simulation.

To create cost prior for experiments in Section~\ref{subsec:mismatch_experiments} we collected 50,000 evaluations of 30s trials for a range of controller parameters. Then we conducted 50 runs, using random subsets of 35,000 evaluations to construct the prior. The numbers were chosen such that this approach used similar amount of computation as our kernel-based approaches. To accommodate GP prior with a large number of points we used a sparse GP construction provided by~\cite{GPMLCode}.